\documentclass{article}
\usepackage{microtype}
\usepackage{graphicx}
\usepackage{subfigure}
\usepackage{booktabs}
\usepackage{hyperref}

\usepackage[accepted]{icml2021}
\usepackage{amsmath}
\usepackage{amsfonts}
\usepackage{bbm}
\usepackage{enumitem}
\usepackage{siunitx}
\DeclareMathOperator*{\argmax}{arg\,max}

\newcommand\mypound{\scalebox{1.1}{\raisebox{0ex}{\#}}}

\icmltitlerunning{When Can Models Learn From Explanations? A Formal Framework for Understanding the Roles of Explanation Data}

\begin{document}

\twocolumn[
\icmltitle{When Can Models Learn From Explanations? \\ A Formal Framework for Understanding the Roles of Explanation Data}

\begin{icmlauthorlist}
\icmlauthor{Peter Hase}{}and 
\icmlauthor{Mohit Bansal}{}\\University of North Carolina at Chapel Hill\\\texttt{\{peter,mbansal\}@cs.unc.edu}\\
\end{icmlauthorlist}

\icmlkeywords{Machine Learning, ICML, Natural Language Explanations, Natural Language Processing}

\vskip 0.3in
]

\begin{abstract}

Many methods now exist for conditioning model outputs on task instructions, retrieved documents, and user-provided explanations and feedback. 
Rather than relying solely on examples of task inputs and outputs, these approaches use valuable additional data for improving model correctness and aligning learned models with human priors.
Meanwhile, a growing body of evidence suggests that some language models can (1) store a large amount of knowledge in their parameters, and (2) perform inference over tasks in textual inputs at test time. 
These results raise the possibility that, for some tasks, humans cannot explain to a model any more about the task than it already knows or could infer on its own.
In this paper, we study the circumstances under which explanations of individual data points can (or cannot) improve modeling performance. 
In order to carefully control important properties of the data and explanations, we introduce a synthetic dataset for experiments, and we also make use of three existing datasets with explanations: e-SNLI, TACRED, and SemEval. 
We first give a formal framework for the available modeling approaches, in which explanation data can be used as model \emph{inputs}, as \emph{targets}, or as a \emph{prior}. After arguing that the most promising role for explanation data is as model inputs, we propose to use a retrieval-based method and show that it solves our synthetic task with accuracies upwards of 95\%, while baselines without explanation data achieve below 65\% accuracy. 
We then identify properties of datasets for which retrieval-based modeling fails.
With the three existing datasets, we find no improvements from explanation retrieval. Drawing on findings from our synthetic task, we suggest that at least one of six preconditions for successful modeling fails to hold with these datasets.\footnote{Our code and data will be made publicly available at: \url{https://github.com/peterbhase/ExplanationRoles}}

\end{abstract}

\section{Introduction}
\label{sec:introduction}

To provide signal for learning, traditional supervised learning algorithms use labels consisting of class IDs or a number in regression settings. Yet training models with data in this form provides the minimum possible supervision for learning a task. Consider how deeply this style of learning contrasts with the way a person can learn a task by getting verbal explanations from someone helping them in addition to just the error signal from their performance. 
Access to such feedback can accelerate learning, resulting in less error-prone behavior, while also aligning the learned behavior with the teacher's prior on what behaviors are good. Since this sort of training should yield efficient and safe outcomes, the contrast between machine and human learning points to natural question: How can we incorporate natural language \emph{explanations} into learning algorithms?

A long line of past work has sought to use explanations, rationales, instructions, and other similar data to improve models. Proposed methods use explanations to constrain or regularize the learned model \cite{zaidan_using_2007, small2011constrained, ba2015predicting, zhang_rationale-augmented_2016, Srivastava2018LearningCF, andreas2018learning, liang2020alice}, to automatically label data for data augmentation \cite{hancock_training_2018, wang_does_2019, awasthi2020learning}, as additional supervision \cite{narang_wt5?!_2020, hase2020leakage, pruthi2020} or intermediate structured variables \cite{camburu_e-snli:_2018, rajani_explain_2019, wiegreffe2020}, and simply as model inputs \cite{rupprecht2018guide, Co-Reyes2019Guiding, zhou2020towards}.

What is surprising about the sheer breadth of approaches in these works is that they all aim to incorporate essentially the same kinds of information. We can describe each of these approaches as trying to augment models with (1) information not available through their inputs or in their parametric knowledge, or (2) a further specification of the task that is informative about which models are good. Improving models in this manner is a natural goal of approaches using explanations, since one purpose of an explanation is to communicate a mental model \cite{doshi-velez_towards_2017, miller2019explanation}. But how do explanations get used as additional targets, as inputs, as regularizers, as structured variables, and as rules for automatic data labeling? Even under a general notion of what an ``explanation" is, e.g. the answer to some \emph{why-question} \cite{miller2019explanation}, this kind of data plays an impressive number of roles. 

Yet there are tasks where explanations do not fulfill these roles effectively, as improvements in performance prove elusive even when thousands of explanations are gathered \cite{narang_wt5?!_2020, hase2020leakage}. In fact, there is reason to think that for some tasks models will not need additional information or further task specification of the kind explanations provide. This is because large language models now (1) store a great amount of knowledge in their parameters \cite{roberts-etal-2020-much, lewis2020retrieval}, and (2) infer tasks at test time from the input itself \cite{radford_language_2019, gpt3, weller-etal-2020-learning}. So in some situations we may not be able to explain to a model more about a task or a data point than it already knows or could infer on its own. What remains unclear, however, is the set of conditions which distinguish situations where explanations will be helpful from those where they will not be helpful in practice or cannot be in principle. 

In this paper, we (1) give an argument for the role of explanations in modeling that helps us understand how explanations have been used in such distinct ways and points us toward suitable methods, and (2) we experimentally study the conditions under which explanations are or are not helpful to models, using a specially designed synthetic task and three existing datasets with explanations given for individual data points. 
The modeling approach we ultimately propose is to perform retrieval over past explanations and provide them as inputs to a model at prediction time (see Sec.~\ref{sec:our_model}), which is the approach we reach following our broader argument in Sec.~\ref{sec:formalizing_the_role}. 
Our synthetic task (described in Sec.~\ref{sec:synthetic}) is designed to have analogous properties to existing real (i.e., human-curated) data, and it is especially useful here as it enables us to test a number of hypotheses that we could not test with existing datasets.

Using RoBERTa as a representative large language model \cite{liu_roberta_2019} and Sentence-BERT as a retrieval model \cite{reimers-gurevych-2019-sentence},
we investigate a number of \textbf{primary research questions}, each given with brief context: 
\begin{enumerate}[itemsep=0pt, wide=0pt, leftmargin=*, after=\strut]
    \item \textbf{RQ1.} Since some models can infer tasks from sequences at test time, providing task information may not be helpful. \emph{When can models solve our synthetic problem by inferring each sequence's task, and when must they be given the task information?}
    \item \textbf{RQ2.} Explanations seen in the past may help with predicting future data points. \emph{Can retrieval of past explanations enable a model to solve our task?}
    \item \textbf{RQ3.} Useful information might be distributed over several explanations. \emph{Can models aggregate information across explanations for better prediction?}
    \item \textbf{RQ4.} We can let pretrained models combine explanations by giving them as textual input, or we can pool extracted feature representations. \emph{What is the best way to compute explanation representations for prediction?}
    \item \textbf{RQ5.} Good explanations pertain to the data point they are given for, but \emph{what makes an explanation relevant across data points? What enables a retrieval model to find relevant explanations for a new data point?}
    \item \textbf{RQ6.} One intuitive use case for explanations is to encourage models to rely on causal features rather than spurious correlations. \emph{Can explanations help models learn to use strong features rather than weak ones?}
    \item \textbf{RQ7.} Here, the training signal for a retrieval model depends on how the classifier uses the explanations the initial retrieval model can provide.
    \emph{How does the co-dependence between classifier and retrieval model influence the viability of joint training?} 
    \item \textbf{RQ8.} After identifying a set of conditions which determine whether retrieval-based modeling can succeed in our synthetic task, we ask: \emph{does retrieval of explanations improve model performance on existing datasets?}
\end{enumerate}

\section{Formalizing the Role of Explanations \\ \hspace{9pt} in Modeling Data}
\label{sec:formalizing_the_role}
\begin{figure}[t]
\centering
 \includegraphics[width=.49\textwidth]{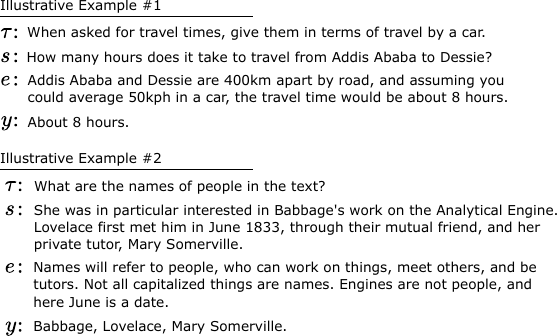}
 \vspace{-15pt}
\caption{Hypothetical data and explanations for illustration. In these examples, $s$ is the kind of input one might expect a model to produce the correct output for after some amount of finetuning on $(s,y)$ pairs. 
For some models $s$ may be sufficient, while others may benefit from additional information as provided by $\tau$ or $e$. 
} 
\label{fig:examples1}
\vspace{-7pt}
\end{figure}

In what follows we discuss what we mean by the term ``explanation" (Sec.~\ref{sec:what_is_explanation}), our formal framework for the uses of explanations in modeling and relevant work on the subject (Sec.~\ref{sec:formal_framework}), a unified view of the roles of explanations in modeling (Sec.~\ref{sec:united_view}), how explanations complement the input in NLP tasks (Sec.~\ref{sec:how_explanations_complement}), and the model we use in this paper (Sec.~\ref{sec:our_model}). 

\subsection{What Is an Explanation?}
\label{sec:what_is_explanation}

The term ``explanation" has no consistent definition in machine learning, as methods papers use it in multiple senses and even opinion papers present definitions of limited specificity. 
For our present purposes, we use the term to refer to the kinds of data one might collect if asking a person to answer the question, ``Why does data point $x$ have label $y$?" This is a generic formulation of the explanation as an answer to a \emph{why-question} of the kind discussed in \citet{miller2019explanation}. For a more extensive discussion of explanations in the context of AI, we refer the reader to this work. Rather than try to give a delimiting, formal definition of the kind of data generated from this question, we proceed with some illustrative examples, shown in Fig.~\ref{fig:examples1}. In Sec.~\ref{sec:experimental_setup}, we describe human explanations used in experiments.

\subsection{Formal Framework and Relevant Work}
\label{sec:formal_framework}
\begin{figure*}
\centering
 \includegraphics[width=.94\textwidth]{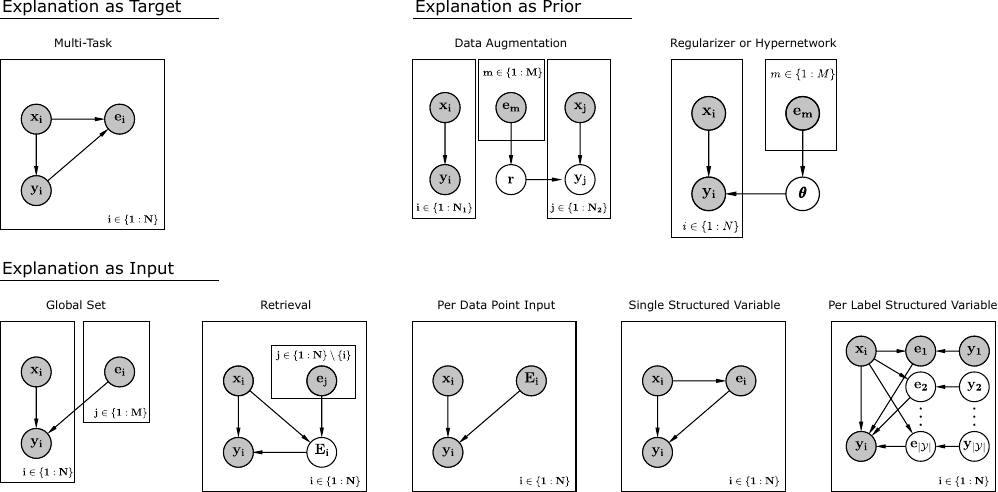}
 \vspace{-2pt}
\caption{Graphical models for several approaches to using explanations as \emph{targets}, as \emph{inputs}, and as \emph{priors}. Typically past works do not condition on human-given explanations at test time, unless they are collected in an interactive manner with a user or specially designed to not leak the data point label. Note prior works may add or remove dependencies in some cases.} 
\label{fig:model_comparison}
\vspace{-7pt}
\end{figure*}

In this section, we lay out our theory of how explanations may be used in modeling a task. With a focus on supervised learning, we characterize the modeling process here in terms of MAP inference over model parameters $\theta$,
\begin{equation*}
    \hat{\theta} = \argmax_\theta p(\theta|X,Y) \hspace{13pt} p(\theta|X,Y) \propto p(Y|X,\theta)p(\theta)
\end{equation*}
where $Y$ is a set of labels for inputs $X$, and the pair constitute a standard supervised learning task. We refer to the role of $Y$ in this probabilistic model as the target, $X$ as an input, and $p(\theta)$ as a prior. Whereas $X$ is intended as the data observed at prediction time, we allow for a latent variable $Z$ to be included as follows:
\begin{equation*}
    p(\theta|X,Y) \propto \int_\mathcal{Z} p(Y|Z,X,\theta)p(Z|X)p(\theta)dZ
\end{equation*}
Both $X$ and $Z$ will be considered as ``model inputs" below. Note that we intend this framework to extend to a many-task situation, which we define as a case where several distinct conditional distributions produce the data $\{Y,X\}$. All that is required is that $Z$ indicate the task $\tau$ to be solved, i.e. which conditional distribution should be computed. 

A few examples: For supervised classification, the task is to map an input $X$ to a label $Y$, and $Z$ could be a document retrieved from a database. In autoregressive modeling, $X$ and $Y$ are sequences of tokens which may appear in either role depending on the current context and positions to be predicted, and $Z$ could be a textual description of the sequence prediction task or a set of unobserved token positions one intends to marginalize over. 

Below we describe existing approaches to using explanations, categorized in our framework. An overview of the corresponding graphical models is shown in Fig.~\ref{fig:model_comparison} in Supplement (\ref{sec:united_view}). We will refer to \emph{tasks} interchangeably with a \emph{function} to be computed and \emph{parameters} of the true model of the data. We mean to index the conditional distribution for each task and refer to the parameterized function that computes it: $p_\tau(y|x)$ = $f_\theta(x)$. 

\paragraph{Using Explanations as Targets.} Explanations may be used as additional supervision (i.e. as $Y$), depending on the ultimate modeling goals (shown as Multi-Task in Fig.~\ref{fig:model_comparison}). For instance, \citet{pruthi2020} consider the use of attention-weight explanations (from a model) as targets in a multi-task framework, and they find that the explanations make for useful targets in helping another ``simulator" model predict the explained model's outputs. Meanwhile, natural language explanations have been treated as targets in a multi-task, sequence-to-sequence framework, using datasets with free-form textual annotations for each data point \cite{camburu_e-snli:_2018, narang_wt5?!_2020, hase2020leakage, wiegreffe2020}. None of these works find improvements in task performance from incorporating explanations. It is surprising and possibly concerning that a model could learn to generate coherent ``explanations" without the learning of this ability influencing the models that are found for the task, as measured by task performance.

\paragraph{Using Explanations as Inputs.} 
Using additional model inputs may be valuable for solving some tasks (i.e. additional $X$ or $Z$). The first family of approaches here uses explanations directly as model inputs for each data point (Per Data Point Input in Fig.~\ref{fig:model_comparison}). \citet{talmor2020teaching} systematically study RoBERTa's ability to combine pieces of knowledge in a reasoning task by including relevant factoids in the text input. In other settings, \citet{Co-Reyes2019Guiding} provide online natural language feedback to RL agents, which helps them learn new tasks on the fly, and \citet{rupprecht2018guide} take a similar approach to interactive image segmentation with language feedback. 

A key question with these approaches is whether it is sensible to collect explanations at prediction time. In an interactive setting, this is reasonable given that human attention is already demanded and system performance is monitored by a human. However, for cases where total automation is a desired outcome, it may not be feasible to collect explanations at test time. There is also a risk of leaking the label through the additional data. Free-form human explanations tend to directly reveal the label when collected for tasks such as NLI and QA \cite{hase2020leakage, wiegreffe2020}. Here, what is essentially the cost of human labeling could be mistaken as an improvement in model performance.

There are a few ways to avoid collecting explanations at test time.
In ExpBERT \cite{murty2020expbert}, a model conditions on vector representations of an input $x$ and a single, ``global" set of explanations in order to make each prediction (shown as Global Set in Fig.~\ref{fig:model_comparison}). 
This can work well for handling up to a hundred or so explanations, but cannot scale to settings with many thousands of explanations.
\citet{zhou2020towards} treat explanations as latent variables when modeling datasets where only a subset of data points have explanations, and at inference time they retrieve explanations from the training data (Retrieval in Fig.~\ref{sec:how_explanations_complement}, with one difference noted here). However, they do not learn the retrieval model, and during training they allow for a data point's own explanation to be conditioned on as its label is predicted. Since explanations are not available for test data points, this leads to distribution shift between training and test-time inference, and it may introduce label leakage during training predictions.
Instead of retrieving explanations, a few works condition on explanations generated at test time using generative models learned with human explanations as supervision, which are represented as Single Structured Variable and Per-Label Structured Variable in Fig.~\ref{fig:model_comparison} \cite{camburu_e-snli:_2018, rajani_explain_2019, kumar_NILE_2020, hase2020leakage, wiegreffe2020}. While this form of intermediate supervision could in principle help models learn useful structured variables (the explanations) for prediction, these methods have not produced sustained improvements in model accuracy.

\paragraph{Using Explanations as Priors.} Here, we group together any approach to defining or learning a distribution over model parameters, including those that condition on some data, $p(\theta|data)$. We note that this is a prior over model weights not in the sense that the distribution is independent of any data (which it is not), but rather in the sense that the posterior parameters are conditioned on the prior. 
One natural way to use explanations is to constrain the learned model, e.g. by constraining the conditional distributions the function can express \cite{Srivastava2018LearningCF, srivastava2018zero}, or through placing priors over how features are weighted or extracted \cite{zaidan_using_2007, small2011constrained, zhang_rationale-augmented_2016, ross2017right, bao-etal-2018-deriving, selvaraju2019taking, liang2020alice, stammer2020right}. Other works map directly from text to parameters in models \cite{ba2015predicting, andreas2018learning}, in effect learning a prior $p(\theta|text)$ (though \citet{andreas2018learning} condition on generated rather than human-provided text at test time). These methods are all effectively described by Regularizer or Hypernetwork in Fig.~\ref{fig:model_comparison}. 
Lastly, a few approaches learn to use explanations for automatically labeling data for data augmentation purposes \cite{hancock_training_2018, wang2019learning, awasthi2020learning}, which is effectively augmenting a task with data drawn from some prior distribution $p_\theta(y|x)$ given by the noisy labeling mechanism (shown as Data Augmentation in Fig.~\ref{fig:model_comparison}). 
Critically, in each of these cases, the prior over model weights is some function of explanations, meaning that we require an \emph{interpretation} $\mathcal{I}$, whether learned or given by humans, of how the explanations encode information about the model. We will write that a prior over models is given by an interpretation function on a set of explanations: $p(\theta|\{e\}) = \mathcal{I}(\{e\}).$ This kind of function can serve either as a regularizer during training or a hypernetwork that directly outputs model parameters or, equivalently, some task representation \cite{ha2016hypernetworks}.

\subsection{How Explanations Achieve One Goal As Targets, Inputs, \emph{or} Priors} 
\label{sec:united_view}
Each of the above methods of supplying information to the modeling process may appear rather distinct, but in principle they can all be used to influence the behavior of a learning algorithm as represented in the posterior parameters.
In fact, we observe situations where a single piece of data can be used either as a target, input, or information yielding a prior. Below, we describe a few such situations in simplified terms, providing some justification for how a single format of explanation data might be used as a label, input, or prior. Ultimately, the fact that these various roles can fulfill a single purpose helps us understand how explanations have historically been used with some success in each of the apparently disparate roles.
We should note that it was already clear that training better models was one \emph{goal} of using explanations in modeling. We would expect a~priori that explanations are suited to this goal given that one underlying purpose to explanation is the communication of a mental model \cite{doshi-velez_towards_2017, miller2019explanation}. 
\paragraph{Using Data as a Target \emph{or} Prior.} Adopting terminology from \citet{pruthi2020}, we refer to a \emph{teacher} giving explanations to a \emph{student} who is learning a task. Suppose a student is modeling a simple 1-D regression problem ($x \in \mathbb{R}$) as $y \sim \mathcal{N}(y \ | \ \theta x, \sigma^2)$, for data $D = \{x_i,y_i\}_{i=1}^n$, using a known $\sigma^2$ and a Normal prior $p(\theta)$. 
In this case, the teacher could in principle induce any MAP estimate they wish by adding a single data point $(x_1, y')$ to $D$, a copy of the first data point with a new label. Of course, the teacher could also induce any desired MAP estimate by directly modifying the student's prior using a particular interpretation function, $p(\theta|y')=\mathcal{I}(y')$. This is simply an illustrative example where one can achieve the same learning outcomes either by providing additional targets or using a particular prior. A more serious analysis would be required to formalize the argument for neural language models and objectives for structured outputs. Thus far, natural language explanations have made no difference to task performance when used as targets \cite{narang_wt5?!_2020, hase2020leakage}. The evidence is more favorable for using attention weights from a model as targets, but \citet{pruthi2020} find this form of explanation to work better as a prior. 

\paragraph{Using Data as an Input \emph{or} Prior.} Now consider a multivariate regression setting with $y\in \mathbb{R}$ and features  $x=(x_1,x_2)$ with $x_1 \in \mathbb{R}$ and $x_2 \in \{0,1\}$, where the true model is: $y$ is linear in a continuous feature $x_1$, with the strength of the relationship modulated by the binary feature~$x_2$ (written as $y=\beta_{11}x_1 + x_2\times\beta_{12}x_1$).
Notice that, per our definition of a task above, $x_2$ is exactly a task representation $\tau$, since it controls for which of multiple functions define a conditional relationship $p(y|x_1)$. 
Hence, we can treat $x_2$ as a task representation and define an interpretation $\mathcal{I}$ to give a prior over the weight on $x_1$, $p(\beta_{1}|x_2) = \mathcal{I}(x_2)$. A model of this form takes the appearance
\begin{align}
    p(y|x) &= p(y|x_1;\beta_{1})p(\beta_{1}|x_2)p(\beta_{1})\\
    \intertext{Interestingly, there will exist equally predictive models of the form (1) as there will for a standard regression model,}
    p(y|x) &= p(y|x_1, x_2; \beta_{11}, \beta_{12})p(\beta_{11}, \beta_{12}).
\end{align}
With the benefit of hindsight, we can say that the simplest interpretation function to represent $p(\beta_1|x_2)$ places a point mass on $\beta_{11}$ when $x_2=0$ and on $\beta_{11}+\beta_{12}$ when $x_2=1$. But we could also learn the prior $p(\beta_1|x_2)$, either with direct supervision for $\beta_1$, by differentiating through a point estimate $\hat{\beta_1}$, or by marginalizing over a random variable for $\beta_1$. In this manner, one can learn equally predictive models treating $x_2$ as an input to a single learned function or a task representation that carries information about model parameters. As before, this is only a simple example, and a more formal analysis would be required to precisely identify this phenomenon when using textual data with methods that may perform interpretation and prediction within one large computation graph (i.e. existing neural models).

\subsection{How Explanations Complement the Input in NLP}
\label{sec:how_explanations_complement}

The ambiguity between considering data as an input or prior is of great relevance in NLP now as a growing body of evidence suggests that pretraining language models teaches them how to do inference over tasks at test time. Indeed it appears that sufficiently large language models do ``infer and perform the tasks demonstrated in natural language sequences in order to better predict them," as \citet{radford_language_2019} hoped for. For example, GPT-3 metalearns how to do sequence prediction over the course of pretraining, which equips it for zero-shot prediction given task descriptions and examples \cite{gpt3}. Even GPT-2 demonstrates the ability to infer the task at prediction time, e.g. for summarization purposes given the ``tl;dr" prompt \cite{radford_language_2019}.

\begin{figure*}
\centering
 \includegraphics[width=.9\textwidth]{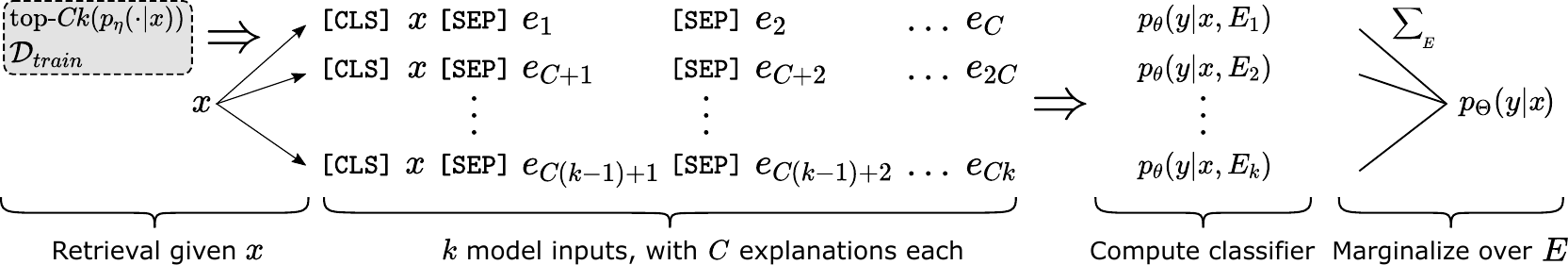}
 \vspace{-4pt}
\caption{A depiction of our retrieval-based method \textsc{\textsc{TextCat}}. A total of $Ck$ explanations are retrieved and allocated into $k$ latent variables, each a set of explanations $E$, which are marginalized over to produce a final prediction.} 
\label{fig:model}
\vspace{-6pt}
\end{figure*}

But these results leave open the question of when and to what degree task information is helpful for prediction in well-defined tasks. In question answering, for example, when should we think that inferring or conditioning on task information is helpful, as opposed to relying on a task's ``input" alone? In fact, for cases like QA, it is even difficult to identify what counts as a \emph{sufficient} input for the task to be solvable by some model without additional information clarifying ambiguities in the task or providing relevant background knowledge. Consider that \citet{roberts-etal-2020-much} use pretrained models to answer questions without any further input, while \citet{lewis2020retrieval} find it helpful to retrieve relevant documents from Wikipedia before answering, drawing a distinction between parametric and non-parametric model memory. Yet when \citet{weller-etal-2020-learning} study how models generalize across tasks when conditioning directly on task descriptions, they formulate the descriptions as \emph{questions} with the task's data given as accompanying documents. \textbf{Hence we see one model's input $x$ used as another model's task description $\tau$, and in both situations additional (possibly retrieved) data can improve task performance}.

Our experiments provide some answers to the remaining question of when and degree to what degree task information is helpful,
and based on our experiments in Sec.~\ref{sec:design_and_results},
we describe conditions for models (1) inferring tasks from the input alone, (2) benefiting from the retrieval of additional information, and (3) being able to learn the retrieval.

\subsection{Assumed Structure of Data}

In this paper we assume we have data of the form $D=\{x_i,y_i,e_i\}_{i=1}^n$, where $(x,y)$ is a standard data point in a supervised classification task, and $e$ is some kind of data collected in response to a question like ``why does data point $x$ have label $y$?" In our experiments with both synthetic and human-curated data, $x$ and $e$ are sequences of tokens from a fixed vocabulary. 
The approaches we present will allow for unexplained training data, meaning some or even most $e_i$ may be missing. The model may use any number of free-floating explanations too, i.e. $e_i$ without corresponding $(x_i,y_i)$ pairs, though this does not apply to datasets in this paper.

\subsection{Our Model}
\label{sec:our_model}

Here, we introduce our chosen model for incorporating explanation data, which makes use of explanations as \emph{model inputs} after they are retrieved from the training data (the ``Retrieval" graphical model in Fig.~\ref{fig:model_comparison}). 
Given our discussion above, a few reasons point us in this direction:
(1) since past explanations may be useful for future predictions, while collecting explanations is costly and can lead to label leakage, we want to avoid collecting explanations at test time; 
(2) this method may directly condition on relevant information that is useful for reasoning tasks \cite{talmor2020teaching}; 
(3) textual data can provide useful task information when serving as a model input, and hence this is a natural way to learn a prior over tasks \cite{gpt3, weller-etal-2020-learning} 
(4) retrieval is more scalable than conditioning on a global set of explanations, and
(5) using explanations as structured variables and as targets do not appear to be promising approaches at the moment \cite{hase2020leakage, wiegreffe2020, pruthi2020}.

So, we use a retrieval-based model that treats retrieved explanations as latent variables to be marginalized over. Our approach is similar to \citet{lewis2020retrieval}, who marginalize over latent documents retrieved from Wikipedia for question answering, question generation, and fact verification. 
The marginal distribution is given as:
\begin{equation*}
    p_\Theta(y|x) = \sum_{e\in \text{top-}k(p_\eta(\cdot|x))}p_\theta(y|x,e)p_\eta(e|x)
\end{equation*}
where top-$k$ gets the top $k$ texts as ranked by the retrieval model, $p_\eta(e|x)$. \textbf{Note that we never retrieve a data point's own explanation when predicting its label}. We do so because explanations can leak the label \cite{hase2020leakage} and this approach matches the test-time distribution, where we assume explanations are not collected for new data points (see discussion in Sec.~\ref{sec:formalizing_the_role}). 

\citet{zhou2020towards} also propose to use explanations as latent variables and retrieve explanations at inference time, but they do not learn the retrieval model, marginalize over the latents during inference, or prohibit data point's own explanations from being retrieved. 
In our experiments, we compare with their original approach and a version where we marginalize over the latents and learn the retrieval model. 

The form of $p_\eta(e|x)$ follows \citet{lewis2020retrieval} and \citet{karpukhin-etal-2020-dense}. Given a query $x$, unnormalized probabilities are computed as:
\begin{equation*}
    p_\eta(e|x) \propto \exp{(f_\eta(e)^Tf_\eta(x))}
\end{equation*}
where $f_\eta$ embeds each sequence into a vector. To compute top-$k(p_\eta(\cdot|x))$, we search through the training explanations using FAISS \cite{JDH17}. We discuss methods for computing $p_\theta(y|x,e)$ and $f_\eta(e|x)$ in Sec.~\ref{sec:methods}.
Because it may be helpful to reason over multiple explanations at once, we extend this model to allow for explanations to be composed into a single ``document." Assuming explanations to be conditionally independent given $x$, we can compute the probability of a set of explanations $E=\{e_c\}_{c=1}^C$ as
\begin{equation*}
    p(E|x) \propto \exp{(\sum_{e\in E}f_\eta(e)^Tf_\eta(x))},
\end{equation*}
where (1) a \emph{context size} $C$ will control the size of the explanation set, (2) a value of $k$ implies that the top $Ck$ will be retrieved, and (3) we sort these $Ck$ explanations into sets in order of their probability $p_\eta(e|x)$.

We represent the overall approach in Fig.~\ref{fig:model} for one method of computing $p_\theta(y|x,E)$ (described fully in Sec.~\ref{sec:methods}), where explanations are concatenated with the query sequence. Flowing from left to right, Fig.~\ref{fig:model} shows how explanations are retrieved from the training data conditioned on a query sequence $x$, then allocated into $k$ classifier inputs with $C$ explanations each. The $k$ classifier predictions are aggregated by marginalizing over the latent variable, $Z=E$. 

\paragraph{Modeling Assumptions.} In using retrieval, we make a few assumptions. First, since the number of forward passes per data point scales with $k$, we require a relatively small value of $k$, i.e. $k\leq10$, for reasonable computational efficiency in SGD-based training. 
Hence, we must assume that this summation is sufficiently similar to the full summation over latent variables. This assumption is more likely to hold when (1) a small number of documents account for most of the probability mass in $p_\eta(e|x)$, and (2) a pretrained model $p_\eta(e|x)$ yields a decent initial rank-ordering, such that some of the best documents are in the top-$k$. The exact value of $k$ we use depends on the experiment. 
A second, more basic assumption is that explanations will be useful in predicting other data points' labels. Such an assumption is needed since we never condition on a data point's own explanation. 
We study how the ``relevance" of explanations to other data points influences task solvability through experiments in Sec.~\ref{sec:rq5}. Lastly, during retrieval we assume that explanations are independent given $x$, i.e. $p(E|x)=\prod_{e \in E}p(e|x)$. This could be a poor assumption when, for instance, explanations each contribute one of a number of needed facts, in which case it would be helpful to retrieve additional explanations conditioned on what has already been retrieved. 

\section{Synthetic Task}
\label{sec:synthetic}

We design a synthetic dataset so that we can carefully control several important properties of the data, though we also make use of several human-curated datasets (described in Sec.~\ref{sec:experimental_setup}).
Designing a synthetic dataset for the task at hand is a useful exercise for a number of reasons. At a high level, it helps us formalize our intuitions regarding what makes the task solvable or not solvable given (1) certain inputs, (2) certain modeling approaches, and (3) certain available explanations. A critical part of this procedure is that, as we do so, we make \emph{disputable decisions} regarding how the synthetic task maps back onto reality. When all is said and done, one can ask if the properties of the proposed data and modeling paradigm do in fact reflect how we expect modeling will work with human-given, natural language explanations. In this spirit, we claim that our synthetic task shares a few important properties with human-curated data, which are described in Sec.~\ref{sec:data_properties}.
Lastly, as a practical matter, it allows us to study how various properties of the data allow for successful modeling with existing methods. In this paper, we are able to provide experimental answers to six of our eight primary research questions only through synthetic data, and not with available datasets. 
Hence, we introduce a synthetic task for our present purpose. For further discussion of the pros and cons of synthetic datasets, see \citet{liu2021can}. 
In Fig.~\ref{fig:examples2}, we show an example data point, along with a description of how it gets its label. The premise of our task is to classify sequences using counts of different integers in the sequences. The basic idea of counting integers is drawn from \citet{de-cao-etal-2020-decisions}. They propose a toy task requiring a model to count whether there are more 8s than 1s in a sequence, with the purpose of evaluating a model interpretation method.

\subsection{Generated Data}
\begin{figure}[t!]
\centering
 \includegraphics[width=.48\textwidth]{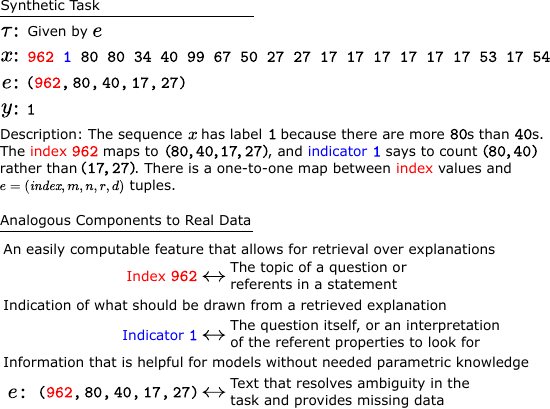}
 \vspace{-16pt}
\caption{Examples of our synthetic task and analogies we draw to human-curated existing data. 
} 
\label{fig:examples2}
\vspace{-8pt}
\end{figure}

We wish to design a task where, while it would be possible to solve the task by learning a function $y=f(x)$, it would be easier if you could condition on relevant explanations and learn $y=f(x,e)$. We propose a few task variants, but the core of the task is that, given a sequence $x$, the binary label will be determined by whether there are more of an integer $m$ in the sequence than there are of an integer $n$. 
We assign a one-to-one map between the integers $(a,b)$ to be counted and a set of special integers each sequence includes as its first two elements, which we term the \textit{index} and \emph{indicator}. 
For our purposes, a key property of this kind of task is that a model could succeed by memorizing the map between (\emph{index}, \emph{indicator}) and the integers it needs to to count. However, it should be much easier to solve the task when directly conditioning on those integers, i.e. learning a function from $(x,a,b)$ to $y$. Here, the ``explanation" $(a,b)$ is a plausible answer to the question of why data point $x$ has label $y$, because this information determines the feature that causes the label. 

Rather than just using the \textit{index} to map to the two numbers that need to be counted, we include the \emph{indicator} so that models can succeed by integrating information from $x$ and $e$. 
An explanation is given as $(\textit{index}, m,n,r,d)$, where either $(m,n)$ or $(r,d)$ is the integer pair that actually needs to be counted. The opposite pair will be a distractor feature whose relative counts match those of the causal feature 50\% of the time. Then the \textit{index} will map to $(m,n,r,d)_{index}$, and the \emph{indicator}, either a $1$ or a $2$, will tell whether it is the first integer pair in the explanation $(m,n)$ or the second $(r,d)$ that needs to be counted (as displayed in Fig.~\ref{fig:examples2}). 
As a result, with \textit{num-tasks} many \textit{index} values, there will be $2 \times \textit{num-tasks}$ possible pairs of integers that have to be counted.
In general, we will refer to a sequence's \emph{task} as the function that counts the relevant integers for that particular sequence, meaning we view our dataset to be composed of many (similar) tasks, each well-defined for a set of sequences.

We next describe the exact dataset in detail. The full generative process is given in Appendix \ref{sec:synthetic_task_generation}. 
We give typical values that dataset parameters take on, and in Sec.~\ref{sec:design_and_results}, we note differences from this default setting as they become relevant to each experiment. The resulting data is:
\begin{enumerate}[itemsep=1pt, wide=0pt, leftmargin=*, after=\strut]
\item Train set: $5000$ sequences of $20$ integers (including \textit{index} and \emph{indicator}), where there are $500$ unique values of \textit{index} in the dataset drawn from $unif(1,10000)$. \textbf{For each \emph{index}, there are 10 distinct $x_i$ that share a common explanation}, (\emph{index}, $m,n,r,d)_{\textrm{\emph{index}}}$. The values of $m, n, r,$ and $d$ are drawn from $unif(1,100)$ while filtering samples s.t. $m {\neq} n {\neq} r {\neq} d$. The corresponding $10$ values of \emph{indicator} are balanced between $1$ and $2$. Half of the points have label $y{=}1$, meaning that either $m{>}n$ or $r{>}d$, depending on which feature is causal. Half the time, the non-causal integer pair in $(m,n,r,d)$ (i.e., the one not indicated by \emph{indicator}), has counts with the same rank-ordering as the causal feature's counts. In each $x_i$, after $m,n,r,$ and $d$ have been randomly placed into the sequence, any unfilled slots are filled with samples from $unif(1,100)$.
\item Dev set: \num[group-separator={,}]{10000} points, none appearing in Train, with the same $500$ \textit{index} values, and twice the number of points per \textit{index} as Train. 
\item Test set: \num[group-separator={,}]{50000} points of similar construction to the Dev set, but with five times the points per \textit{index} as Train.  
\end{enumerate}

\subsection{Important Data Properties}
\label{sec:data_properties}

\paragraph{Analogous Properties to Human-Curated Data.} We claim that aspects of our synthetic task are analogous to properties real (i.e. existing, human-curated) data might take on. 
We first highlight a few properties of the Illustrative Examples in Fig.~\ref{fig:examples1}. Here, $s$ is the kind of input for which one might expect a model to produce the correct output after some amount of finetuning on an appropriate dataset, while $\tau$ offers explicit task instructions and $e$ is an explanation of the data point's label. 
We expect that, for some models, $\tau$ and $e$ will provide useful additional information for the task that is not represented in $s$ or is difficult to infer from $s$. Models might more easily extract this information from $\tau$ or $e$ than they can infer it from $s$, allowing for better task performance. However, a model may infer any ``hidden" information perfectly well without relying on these variables, especially after some amount of finetuning on $(s,y)$ pairs. Without finetuning, a model may already be pretrained to interpret task instructions \cite{gpt3}, or the model may already know the hidden information \cite{roberts-etal-2020-much}, meaning the knowledge encoded is in their parameters and accessible in the right circumstances.

Now, regarding our synthetic data, we first claim that $e$ is an explanation in the sense that it is a plausible answer to the question, "why does point $x$ have label $y$?" The explanation gives the information which determines the feature that causes the label, i.e. the integers that should be counted. We suggest that the \textit{index} in a sequence is analogous to the topic of a question or the referents of a statement (the things referred to): both are computable features that make retrieval-based modeling possible. Likewise, good models will combine the \emph{indicator} and explanation to identify the causal feature in the same way that a good QA model would figure out what to look for in a document by first understanding what the question asks for or the referent properties it should be looking for. 

Our task shares another important characteristic with human-curated data: whenever retrieval could be helpful, models can learn to directly infer the hidden information from the input alone. In the synthetic task, this looks like learning the function from the \emph{index} to the integers to be counted. With question answering, for example, a model could learn the map between a certain topic and the set of facts that could be needed to answer questions about that topic. This may be harder than learning a retrieval model for a given dataset, but it is possible in theory and would render the additional data for retrieval irrelevant. In our experiments in Sec~\ref{sec:rq1}, we outline situations where this map is learned by models, making retrieval unnecessary.

\paragraph{Data Parameters, \emph{Relevance}, and Strong Features.} 
There are a few parameters to the data generation that heavily shape our expectations of the task's solvability. The first is the number of unique values of \textit{index}, which we will refer to as the number of tasks, \textit{num-tasks}. With a fixed training set size, \textit{num-tasks} determines the number of data points per task, $n_\emph{task}$. For example, while we will typically have $10$ points per task, decreasing the number of tasks to $100$ would mean there would be $50$ points per task (with $5000$ training points). This is a particularly important property because it determines how explanations will be \emph{relevant} across data points. Here, we define an explanation for one data point $e_i$ to be \emph{relevant} to another sequence $s_j$ when $e_i$ is informative about what sequence $s_j$'s task $\tau_j$ is. Recall that by $\tau_j$ we refer to the function counting the integers $(a,b)_j$. Formally we will say that a \emph{relevance function} on $s$ and $e$ yields some distribution over the task parameters:
\begin{align*}
    p((a,b)_j|s_j,e_i) = f(s_j,e_i).
\end{align*}
In the standard version of our synthetic task, one such relevance function could place all probability mass on $(m,n)$ if $indicator=1$ and the \textit{index} in $s_j$ and $e_i$ matched (or $(r,d)$ if $indicator=2$). If the \textit{index} does not match, then there would be no information about what $\tau_j$ is, since we randomly sample \textit{index} and $(m,n,r,d)$ values when pairing them. To obtain a smoother, more continuous level of relevance between sequences and explanations, we can also define a predictable relationship between \emph{index}$_i$ and $(m,n,r,d)_i$ so that $(a,b)_i$ and $(a,b)_j$ are close together (under some distance metric) whenever \emph{index}$_i$ and \emph{index}$_j$ are close together. We describe experiments comparing the two settings of binary and smooth relevance in Sec.~\ref{sec:rq5}.

Next, note that we can vary the degree to which the non-causal feature is correlated with the causal (strong) feature. In the case of perfect correlation, we have that $\mypound m{>}\mypound n$ iff $\mypound r{>}\mypound d$ and $\mypound m{<}\mypound n$ iff $\mypound r{<}\mypound d$, regardless of which is the causal feature. This allows us to test whether explanations can induce models to rely on causal rather than non-causal (weak) features. While this is an intuitive reason for thinking explanations should be helpful for models, we show in Sec.~\ref{sec:rq6} that models can correctly use explanations for selection between correlated features only in a narrow set of situations. 

Finally, \textit{index} can be removed from each sequence to more closely imitate a situation requiring task inference. While in principle models can learn the map from \emph{(index, indicator)} to $(a,b)$, in fact we find that models will infer the task even when \textit{index} is removed from the sequence (Sec.~\ref{sec:rq1}). Ostensibly they do so by counting the sequence integers: those which appear often are likely to make up $(m,n,r,d)$. 

\subsection{Kinds of Explanations}
\label{sec:kinds_of_exps}

The data we have described so far includes only a single form of explanation, $e=(\emph{index},m,n,r,d)$, which we will call our \textit{full-info} condition. As long as a retrieval model returns relevant explanations, the task for a sequence can be read off from this kind of explanation. Yet, rather than giving a full description of the task, explanations in existing datasets tend to only partially specify a task or give just a piece of the hidden information for a data point, especially when annotators limit the length of their explanations to a single sentence \cite{camburu_e-snli:_2018, wang2019learning}. 

This leads us to suggest two alternative forms of explanation in our synthetic task, which we refer to as \emph{evidential} and \emph{recomposable} explanations. Given an \textit{index}, \mbox{\textbf{evidential}} explanations are generated by adding independent, zero-mean noise to each element in the true $(m,n,r,d)_\emph{index}$, s.t. taking the average across a set of evidential explanations converges in the limit to the true $(m,n,r,d)_\emph{index}$. In our experiments, we will add some noise $\epsilon$ drawn from the uniform discrete distribution from $-2$ to $2$. 

The second explanation kind, \textbf{recomposable}, is designed so that one could infer the task if one had all the relevant explanations for a particular \textit{index}. We create such a situation by breaking the $(m,n,r,d)$ into parts that neatly recompose back into the true set of numbers. Principally, we do so by dividing the explanation into two pieces, $(m,0,r,0)$ and $(0,n,0,d)$, where some points with that \textit{index} have one explanation, and other points have the other. We ensure that both pieces of an explanation appear at least once among the data points for each \emph{index}. We also experiment with a similar setting where we decompose explanations into four pieces, but do we not include results for this condition as we find them to be quite similar to the two-piece setting.

\section{Computational Methods}
\label{sec:methods}

In this section we describe the methods used to compute $p_\theta(y|x,E)$ and $p_\eta(e|x)$ (see Sec.~\ref{sec:our_model} for the overall model description). For the classifier $p_\theta(y|x,E)$, we use two methods, \textsc{\textsc{TextCat}} and \textsc{H-Mean}, which are described below. Then we describe the retrieval model, which is based on Sentence-BERT \cite{reimers-gurevych-2019-sentence}.

\subsection{Conditioning Mechanisms}
\label{sec:conditioning_mechanisms}

\paragraph{\textsc{\textsc{TextCat}}.} Represented in Figure \ref{fig:model}, this method takes a straightforward approach to conditioning on a set of explanations: concatenating $C$ explanations and the input $x$ to form a longer sequence of text. Each of the original sequences is separated by a special token, e.g. \texttt{[SEP]} for BERT. 
In our experiments, we pass this longer sequence into a RoBERTa-base model. After pooling the output token representations, we pass the resulting vector to a 1-layer MLP for classification. We use mean pooling for our synthetic task and NLI; for relation extraction tasks, we concatenate the representations corresponding to the initial tokens in the \emph{subject} and \emph{object} words, since this is an especially effective pooling technique~ \cite{baldini-soares-etal-2019-matching}.

This approach allows the model to reason over all of the explanations and the input together. While the method may be limited by the fact that some models can face difficulties in processing long pieces of text \cite{beltagy2020longformer}, this issue is partly mitigated by marginalizing over $k$ sets of explanations. As a result of the marginalization, the final prediction can be conditioned on a far higher number ($Ck$) of individual explanations than could fit in the context alone. 

\paragraph{\textsc{H-Mean}.} By \textsc{H-Mean}, we refer to the kind of unweighted hidden representation averaging used in \citet{Co-Reyes2019Guiding} and \citet{zhou2020towards}. \textsc{H-Mean} works by first obtaining representations of the input $x$ and a single explanation $e$ at a time, then passing the unweighted average of these representations to an MLP. For a fair comparison with \textsc{\textsc{TextCat}}, we use the same token pooling and a 1-layer MLP. So with $C$ explanations to condition on, $x' = concatenate(x,e)$, and vector representations from $\textrm{RoBERTa}(x')$, \textsc{H-Mean} obtains a single representation as
\begin{align*}
    h=\frac{1}{C}\sum_{c=1}^{C} \textrm{RoBERTa}(x')
\end{align*}
which is then passed to the MLP for classification. 
\mbox{\textsc{H-Mean}} does not face the same sequence length limitations as \textsc{\textsc{TextCat}}, but by separately processing of each explanations \textsc{H-Mean} may fail to integrate information across explanations. This method also becomes expensive when we marginalize over $E$ (which is what allows retrieval to be learned), as it requires $Ck$ forward passes for a single prediction. We compare the two methods in Sec.~\ref{sec:rq4}.

\subsection{Retrieval} 

We use a similar approach to retrieval as in \citet{lewis2020retrieval}, namely using vector representations of sequences from a pretrained transformer to compute
\begin{equation*}
    p_\eta(e|x) \propto \exp{(f_\eta(e)^Tf_\eta(x))},
\end{equation*}
which is followed by computing top-$Ck(p_\eta(\cdot|x)$. We use an approximate but sub-linear time search method (FAISS) to find the top-$Ck$ points \cite{JDH17}. In our experiments we find that it is necessary to use Sentence-BERT \cite{reimers-gurevych-2019-sentence} as our pretrained $f_\eta$, rather than simply a pretrained RoBERTa model (discussed in Sec.~\ref{sec:rq7}). Sentence-BERT is a network trained to produce semantic representations of sentences that can be compared under cosine similarity. In our experiments, we use the Sentence-RoBERTa-base model trained on a combination of several NLI and semantic textual similarity tasks, with mean pooling of token representations. We normalize the representations we obtain from this model, so that our inner product is equivalent to a cosine similarity.  

Note that during training, we never condition on a data point's own explanation when predicting its label.
This is an important constraint for matching the train and test-time distributions. 
At test time, we assume we have access only to past (training) explanations, since they can be expensive to collect and conditioning on explanations at test time can lead to label leakage, meaning what is essentially the benefit of human labeling could be mistaken as improvements in model performance.

\section{Experimental Setup}
\label{sec:experimental_setup}
Here, we detail the datasets and important model training details used in our experiments.

\paragraph{Datasets.}
\label{sec:real_datasets}

\begin{table}[t]
    \centering
    \small
\begin{tabular}{l l l l l l} 
\toprule
& \multicolumn{4}{c}{Sample Size} \\
\cmidrule(lr){2-5}
Dataset & Explns & Train & Dev & Test & $|\mathcal{Y}|$ \\ 
\midrule
Synthetic   & 5000       & 5000        & 10000   & 50000 & 2\\
e-SNLI      & 549,367    & 549,367     & 9842    & 9824 & 3\\
SemEval     & 203        & 7016        & 800     & 2715 & 19\\
TACRED      & 169        & 68,124      & 22,631  & 15,509 & 42\\
\bottomrule
 \end{tabular}
 \vspace{-5pt}
\caption{Statistics for Datasets.}
\label{tab:data_statistics} 
\vspace{-7pt}
\end{table}

The standard version of our synthetic task used in experiments is described in Sec.~\ref{sec:synthetic}. We include experiments with three other (English) datasets. The first, e-SNLI, is the SNLI dataset annotated with human explanations \cite{bowman_large_2015, camburu_e-snli:_2018}. The next two, SemEval and TACRED \cite{hendrickx-etal-2010-semeval, zhang2017tacred}, are relation extraction tasks with a subset of data points annotated by \citet{wang2019learning}. Summary statistics from the three datasets are shown in Table~\ref{tab:data_statistics}. 
For additional details including data preprocessing see Appendix~\ref{sec:app_experimental_details}.

\paragraph{Model Training.}
\label{sec:model_training}

\begin{table}[t]
\begin{center}
\small
\begin{tabular}{p{0.01\textwidth} p{0.4\textwidth}}
\toprule 
\addlinespace[4pt]
\mbox{\hspace{-5pt}e-SNLI} & \\
\addlinespace[1pt]
$x:$ & \emph{Premise:} After playing with her other toys, the baby decides that the guitar seems fun to play with as well. \emph{Hypothesis:} A blonde baby. \\
$y:$ & Neutral \\
$e:$ & Not all babies are blonde. \\
\midrule
\addlinespace[4pt]
\mbox{\hspace{-5pt}SemEval} & \\
\addlinespace[1pt]
$x:$ & The SUBJ originates from an OBJ which transcends the speaker. \\
$y:$ & Entity-Origin \\
$e:$ & The phrase ''originates from an" occurs between SUBJ and OBJ and there are no more than four words between SUBJ and OBJ and OBJ follows SUBJ. \\
\midrule
\addlinespace[4pt]
\mbox{\hspace{-5pt}TACRED} & \\
\addlinespace[1pt]
$x:$ & SUBJ's husband OBJ died in 1995. \\
$y:$ & Person-Spouse \\
$e:$ & Between SUBJ and OBJ the phrase ``'s husband" occurs and there are no more than five words between SUBJ and OBJ. \\
\bottomrule
\end{tabular}
\end{center}
\vspace{-5pt}
\caption{Example data points from the three existing datasets. More examples can be found in Table~\ref{tab:more_data_examples}.}
\vspace{-7pt}
\label{tab:data_examples}
\end{table}

We train all models in an end-to-end manner using AdamW with a standard cross-entropy loss \cite{loshchilov_decoupled_2017}. This would be straightforward given the model's end-to-end structure, except for the fact that with after every gradient update, \emph{all} training explanation representations need to be recomputed in order for future predictions and gradients to reflect the new parameters. Prior work using retrieval models has either periodically updated the document representations \cite{guu2020retrieval} or left them fixed and only updated the query embeddings \cite{lewis2020retrieval}. We find it is important to update all embeddings at least every epoch, and unless otherwise noted we rebuild the embeddings every 20\% of each epoch (see Appendix~\ref{sec:app_training} for further discussion). 

We give important hyperparameters such as the context size $C$ and retrieval parameter $k$ in each experiment description in Sec.~\ref{sec:design_and_results}. We provide an analysis of the influence of hyperparameters on training in Appendix~\ref{sec:app_training}, but usually we observe that larger values of $C$ and $k$ yield higher accuracies with more stable training behavior. Other hyperparameters for training are also given in Appendix~\ref{sec:app_training}.

\paragraph{Model Selection and Hypothesis Testing.}

We report and visualize results on our synthetic dataset with confidence intervals representing \emph{seed variance}, which accounts for variability across sampled datasets and random model training behavior. We do not estimate sample variance because it is quite small using a test set of \num[group-separator={,}]{50000} points, with a 95\% confidence interval of e.g. $\pm 0.26$ for a model accuracy of 90\%. Seed variance is estimated from 5-10 random seeds, depending on the condition. See Appendix~\ref{sec:app_experimental_details} for further details of seed variance estimation. In synthetic data experiments, we comment on effects far larger than the confidence intervals and do not conduct hypothesis tests. 

With the three existing datasets, for the majority of conditions, we run three model seeds and select the best model by dev set accuracy. We run only one seed for conditions using the full TACRED training set and the e-SNLI dataset with at least \num[group-separator={,}]{50000} training points. With the selected model, we conduct hypothesis tests for a difference in binomial means to check for differences in test set accuracy.

\section{Experiment Design and Results}
\label{sec:design_and_results}

Below, we give the experimental design and results for each research question in Sec.~\ref{sec:introduction}. The first seven research questions are best answered with our synthetic task, and so they each make use of synthetic data (introduced in Sec.~\ref{sec:synthetic}). See Sec.~\ref{sec:rq8} for results with the three existing datasets.

\subsection{RQ1: When can models solve our synthetic problem by inferring each sequence's task, and when must they be given the task information?}
\label{sec:rq1}
\begin{figure}
\centering
 \includegraphics[width=.48\textwidth]{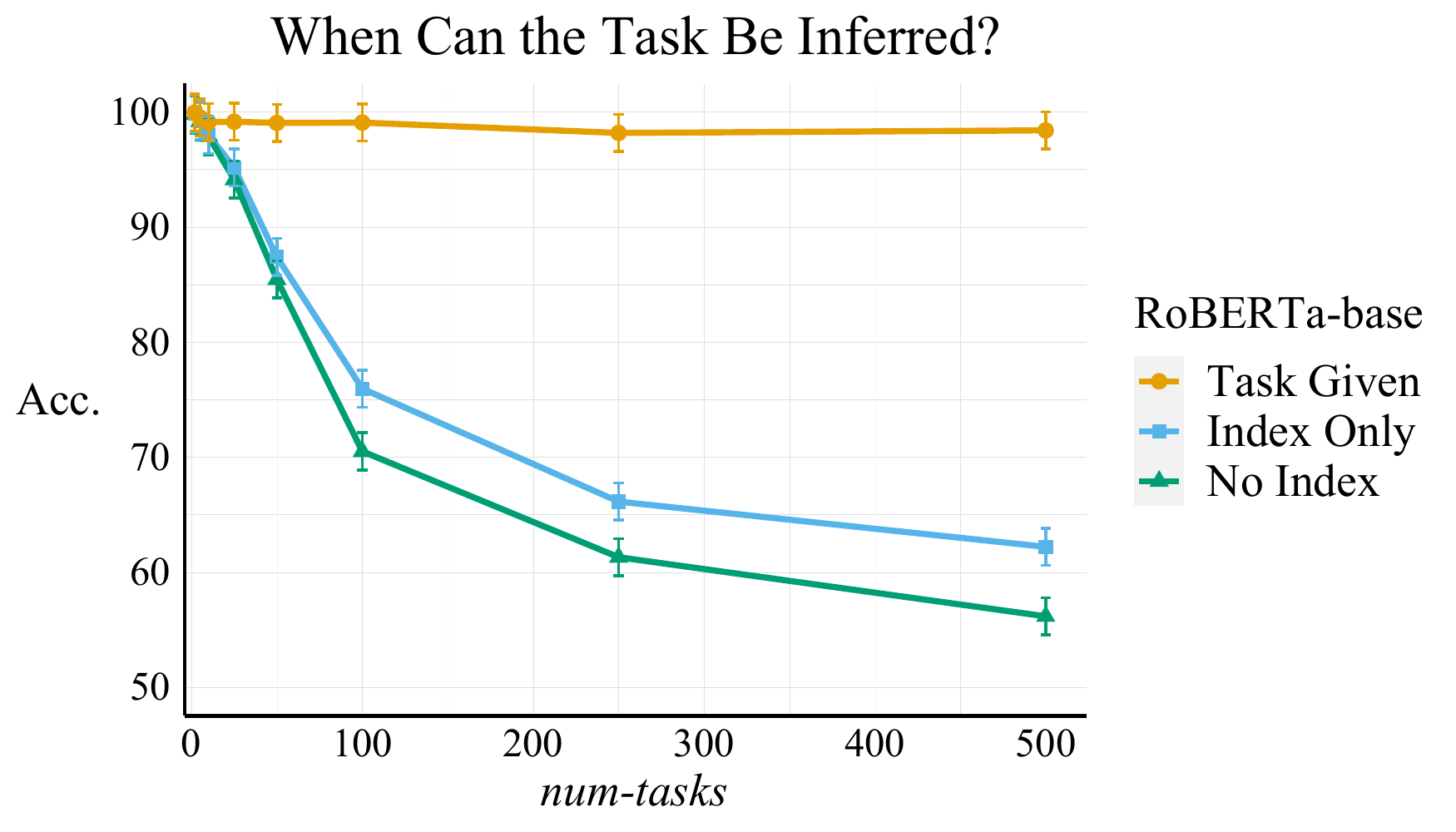}
 \vspace{-16pt}
\caption{(\textbf{RQ1}) Synthetic task accuracy as a function of \textit{num-tasks}.} 
\label{fig:rq1}
\vspace{-8pt}
\end{figure}

\paragraph{Design.} We measure test accuracy as a function of the \textit{num-tasks} parameter across three conditions.
The conditions vary in how task information is available in the input: (1) \emph{task given}, where each sequence has its true task information $(m,n,r,d)$ appended to it; (2) \emph{task signalled}, meaning \textit{index} is given and hence the model can learn the map $index \rightarrow (m,n,r,d)$; (3) \emph{task inferred}, where \textit{index} is not given, so the model must infer the task from the sequence's contents alone. 
To see the interaction between these conditions and model capacity, we test with both RoBERTa-base and RoBERTa-large, and we also measure the effect of increasing the training set size.
Note that, with a fixed training set size, \textit{num-tasks} directly implies the number of points per task, $n_\emph{task}$. In this experiment, \emph{num-tasks}$= \{2,5,10,25,100,250,500\} \Rightarrow n_\emph{task}=\{2500, 1000, 500, 200, 50, 20, 10\}$.

\paragraph{Results.} We show the results in Fig.~\ref{fig:rq1}. We see that, when the numbers of tasks is small, RoBERTa-base can infer the task for each sequence and achieve as high an accuracy as if it had been given task information. Yet, \textbf{the feasibility of task inference quickly falls off as the number of tasks increases} (equivalent to the number of points per task decreasing), reaching accuracies as low as 62.2\% at \textit{num-tasks}$=500$. Meanwhile, we observe that providing the \textit{index} does slightly ease the task inference, but the models can by no means memorize the map from \textit{index} to the task information. Regarding model capacity, we find that using RoBERTa-large increases model accuracy when the number of \textit{num-tasks} is relatively low (less than 250), but after this point RoBERTa-base performs better (see Fig.~\ref{fig:rq1b} in Appendix~\ref{sec:app_experimental_details}). Lastly, we see that increasing the training set size can greatly improve model performance even with \emph{num-tasks}${=}500$, reaching 87.11\% with \num[group-separator={,}]{50000} training points (trend shown in Fig.~\ref{fig:rq1c} in Appendix~\ref{sec:app_experimental_details}). However, we will see in the next section that, in terms of improving model accuracy, even this $10$x increase in training size is less effective than using retrieved explanations with 5000 training points. 

\subsection{RQ2: Can retrieval of past explanations enable a model to solve our task?}
\label{sec:rq2}
\begin{figure}
\centering
 \includegraphics[width=.48\textwidth]{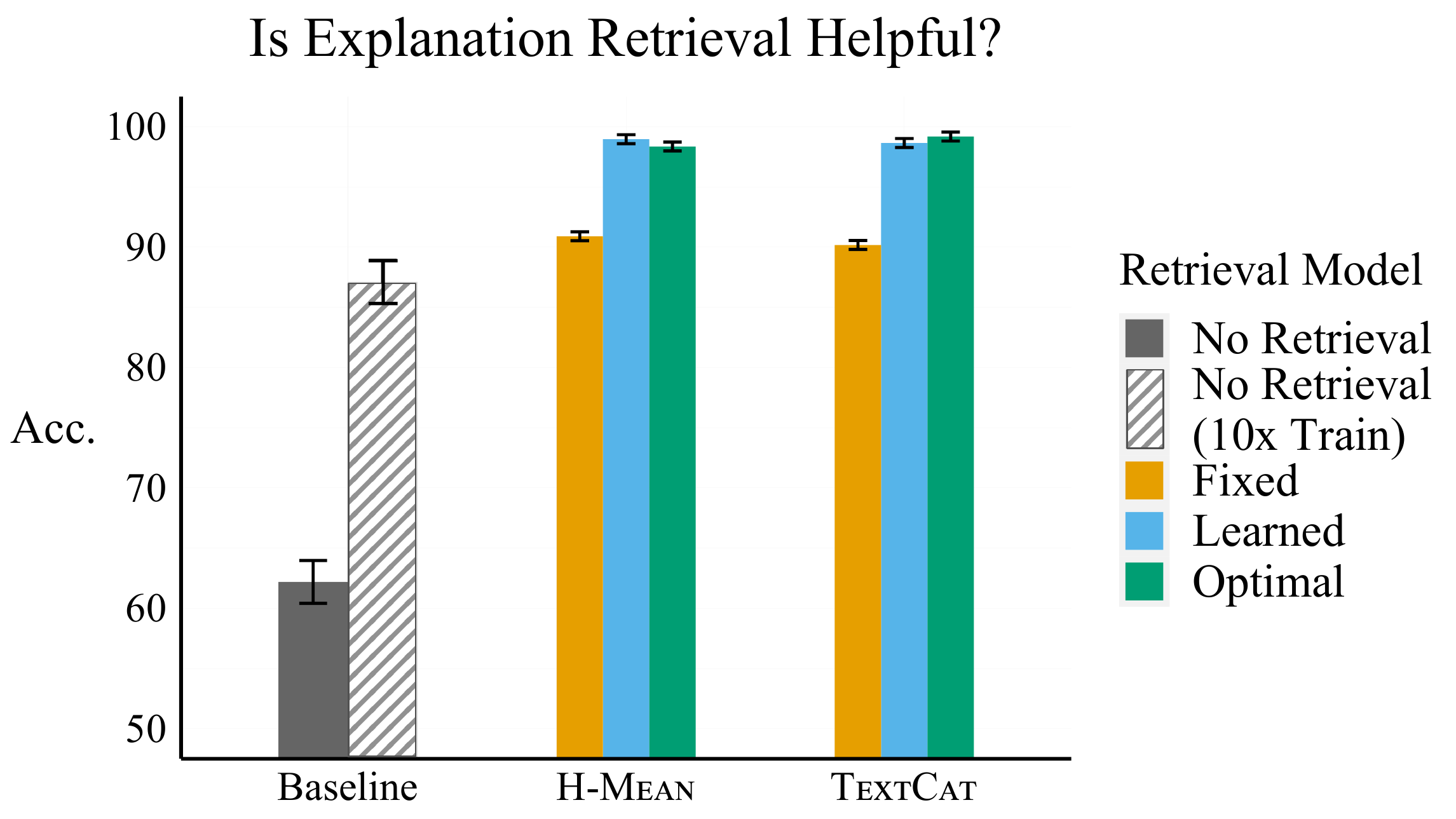}
 \vspace{-16pt}
\caption{(\textbf{RQ2}) Synthetic task accuracy by the conditioning mechanism and retrieval model status, for data with $\textit{num-tasks}=500$.} 
\label{fig:rq2}
\vspace{-8pt}
\end{figure}

\paragraph{Design.} Using the \textit{full-info} explanations and data with \textit{num-tasks}$=500$, we measure model accuracy with retrieval in a $3 \times 2$ design. There are three conditions for the retrieval model: (1) \emph{fixed}, where the Sentence-RoBERTa retriever is fixed and only the classifier is trained, (2) \emph{learned}, where both classifier and retriever are trained end-to-end, and (3) \emph{optimal} where the optimal retrieval model is used and the classifier is trained. Note that we know the optimal retrieval model assigns the highest probabilities to explanations with $\emph{index}_e$ matching the query point's $\emph{index}_x$, so by using a retriever $p(e_i|x_i) = \exp{(\mathbbm{1}[\emph{index}_e=\emph{index}_x])}$ and a context size lower than $n_\emph{task}$, we can ensure the retrieved explanations are always relevant. There are two conditions for the conditioning mechanism used: (1) \textsc{\textsc{TextCat}} with $C{=}k{=}6$, and (2) \textsc{H-Mean} with $C{=}4$ and $k{=}4$, which approximately matches the computational cost of the \textsc{\textsc{TextCat}} condition. 

\paragraph{Results.} Shown in Fig.~\ref{fig:rq2}, the results show that retrieval with Sentence-BERT can reach accuracies above 98\%, improving model accuracy by around 37 percentage points over a no-retrieval baseline. Each conditioning mechanism sees roughly the same improvement. Additionally, we find that the learned retrieval model does as well as the optimal retrieval model, improving over the \textit{fixed} condition by about 7 points. Thus, \textbf{retrieval of explanations allows the model to perform much better than a no-retrieval baseline}. We see a large improvement in performance from retrieval even when the baseline could learn to infer the task information directly from the \emph{index} value in each input. In fact, explanation retrieval outperforms a no-retrieval baseline with as many as \num[group-separator={,}]{50000} training data points (a $10$x increase), which obtains 87.11\% accuracy.  

\subsection{RQ3: Can models aggregate information across explanations for better prediction?}
\label{sec:rq3}
\begin{figure}[t]
\centering
 \includegraphics[width=.48\textwidth]{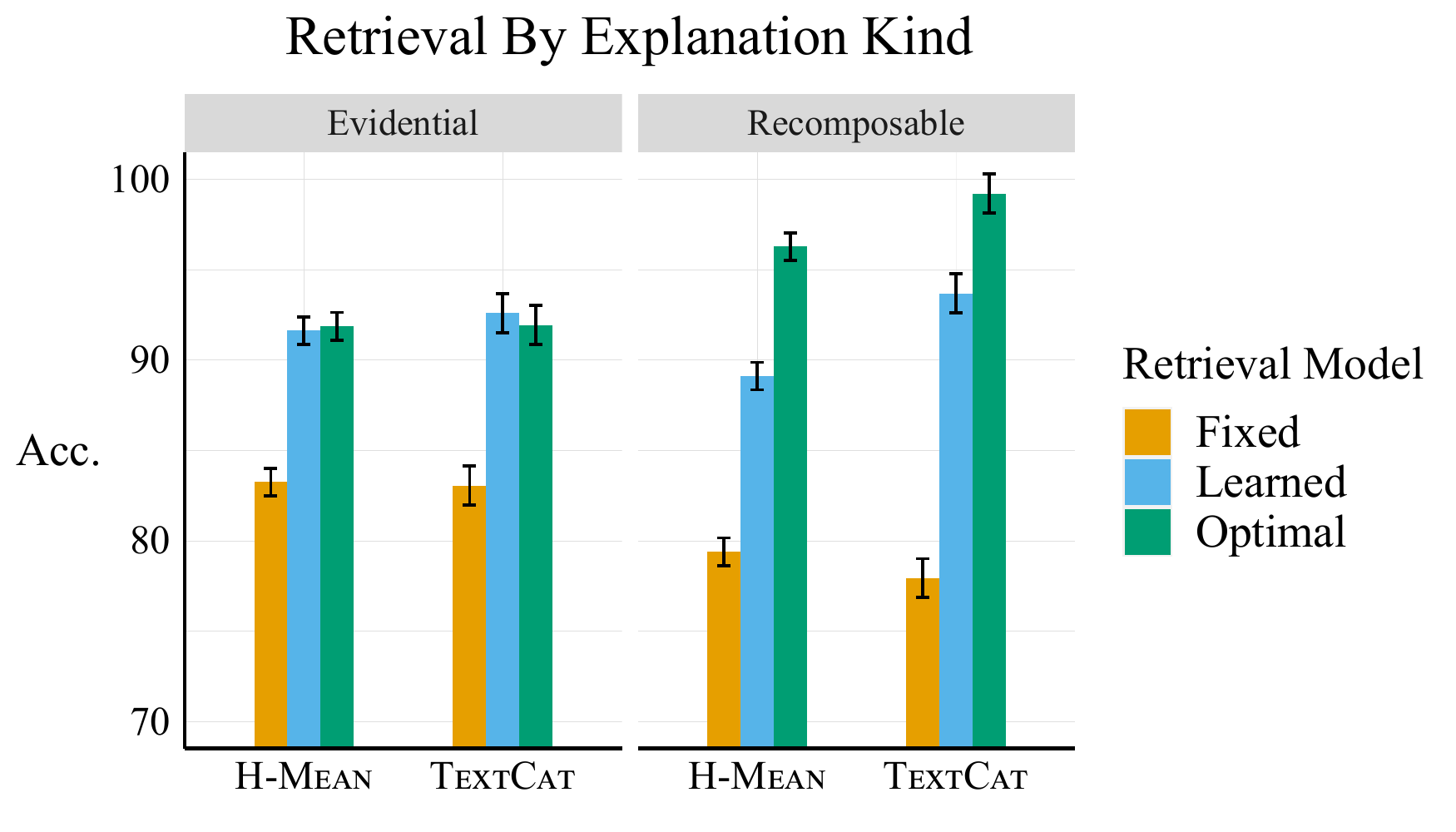}
 \vspace{-16pt}
\caption{(\textbf{RQ3}) Synthetic task accuracy with \emph{evidential} and \emph{recomposable} explanations, grouped by the conditioning mechanism and status of retrieval model.} 
\label{fig:rq3}
\vspace{-8pt}
\end{figure}

\paragraph{Design.} We run the same experiment design as for RQ2, using \emph{evidential} and \emph{recomposable} explanations (see Sec.~\ref{sec:kinds_of_exps}). With evidential explanations, we shift each integer in the explanation (excluding the \emph{index}) independently by zero-mean, discrete noise $\epsilon \sim unif(-2,2)$. In the recomposable setting, for each \emph{index} two explanations combine to give the full task information. As in RQ1, we show results here for values of $C{=}k{=}6$ for \textsc{TextCat} and $C{=}k{=}4$ for \textsc{H-Mean}. 

\paragraph{Results.} We display the results in Fig.~\ref{fig:rq3}. \textbf{With both explanation kinds, the model can learn to retrieve and aggregate information across explanations}, achieving accuracies above 90\%. We observe that for evidential explanations, learned retrieval is close to the optimal retrieval, and the conditioning mechanisms perform very similarly. Yet the models cannot interpret evidential explanations as well as full-info, seeing as even with optimal retrieval both \textsc{TextCat} and \textsc{H-Mean} obtain  around 92\% accuracy compared to full-info's 98\%.

With recomposable explanations, meanwhile, we notice two differences. First, we find that with optimal retrieval \textsc{TextCat} can interpret the recomposable explanations as well as full-info, achieving upwards of 98\% accuracy. Yet we observe that learned retrieval falls 6-8 points short of optimal retrieval (depending on the conditioning mechanism). There is no clear reason why this should be, though we can attribute it to the differences in explanations alone.
Second, \textsc{TextCat} with learned or optimal retrieval outperforms \textsc{H-Mean} with the same retrieval (by 4.58 points for Learned). We discuss this further in the next section. 

\subsection{RQ4: What is the best way to compute explanation representations for prediction?}
\label{sec:rq4}

\paragraph{Design.} Here we rely on results from the experiments for RQ3, and we also test method performance across training set sizes in $\{1000, 1500,2500, 5000, 10000\}$, using optimal retrieval with $C{=}5$ and $k{=}1$ for both \textsc{TextCat} and \textsc{H-Mean}. Lastly, we consider training time as a relevant~factor.

\paragraph{Results.} As shown in Fig.~\ref{fig:rq3}, with learned retrieval \textsc{TextCat} outperforms \textsc{H-Mean} by 4.58 points when explanations are broken down into parts that can be recombined to obtain the exact task information. This is especially important as explanations for existing natural language data can give facts and task specifications that may be combined to form a fuller picture of the problem.
Additionally, we find that for small sample sizes, \textsc{TextCat} achieves higher accuracy than \textsc{H-Mean}, by 9.3 points for $n=1000$ and 9.2 points for $n=1500$, though the gap shrinks to 1.3 points at $n=2500$ and the methods perform equally well after $n=5000$ (see Fig.~\ref{fig:rq4} in Appendix \ref{sec:app_experimental_details}).
As a final consideration, we note that at $C{=}k{=}4$, \textsc{H-Mean} takes 61\% longer to train than \textsc{TextCat} due to the additional model forward passes. 
So, given favorable performance with recomposable explanations and low sample sizes, as well as the training speed, \textbf{\textsc{TextCat} appears to be the preferable conditioning mechanism to \textsc{H-Mean}}, and unless otherwise stated we use \textsc{TextCat} in experiments henceforth. 

\subsection{RQ5: What makes an explanation relevant across data points? What enables a retrieval model to find relevant explanations for a new data point?} 
\label{sec:rq5}
\begin{figure}
\centering
 \includegraphics[width=.48\textwidth]{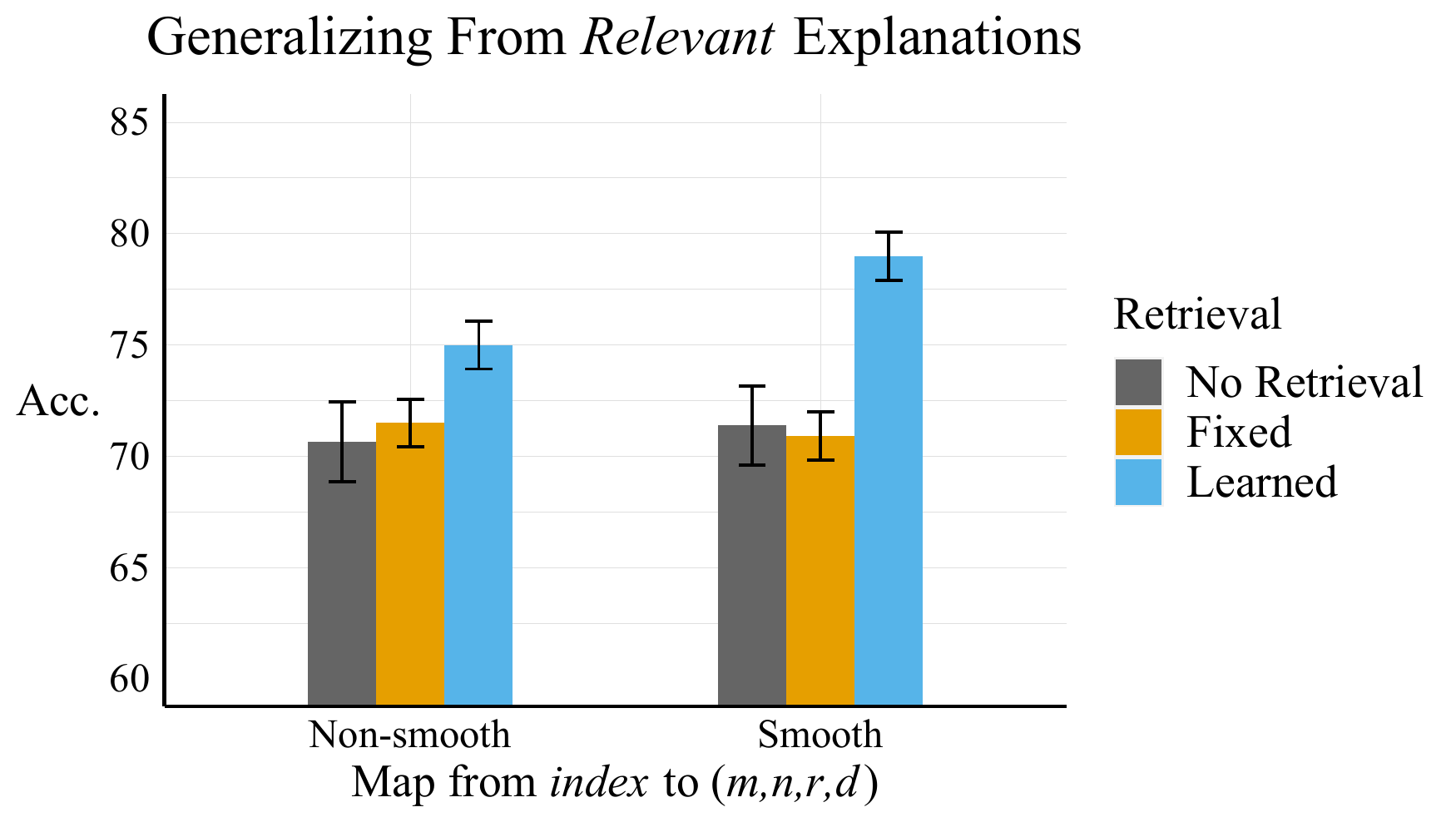}
 \vspace{-16pt}
\caption{(\textbf{RQ5}) Task accuracy with by retrieval model and the \emph{smoothness} of the \emph{index} $\rightarrow (m,n,r,d)$ map, using 1 point per task \emph{index}. At test time new \emph{index} values are used, meaning models must generalize based on retrieved explanations with similar but never exactly correct $(m,n,r,d)$ values.} 
\label{fig:rq5}
\vspace{-8pt}
\end{figure}

\paragraph{Design.} For retrieval-based modeling to be successful, explanations for one data point must be \emph{relevant} to predicting other data points' labels. So far, we have used $n_\emph{task}{=}10$ points per \emph{index}, with test \emph{index} values that have been seen during training, meaning that both during training and testing, ``exactly correct" explanations are available for retrieval (i.e. explanations with the true $(m,n,r,d)$ for the data point at hand). To see what is required for explanations to be relevant across data points, we set $n_\emph{task}$ to 1, making every explanation n the train set unique, and we use test data with \emph{index} values not seen in training. 
As a result of these changes, at both training and test time there are no exactly correct explanations available for retrieval (since we do not retrieve the data point's own explanation). 
In addition, we restrict the causal feature to always be $(m,n)$, rather than $(r,d)$, for reasons that will become apparent momentarily.
To succeed in this setting, models must generalize from explanations given for one data point to a data point with a similar but not identical set of integers to be counted. 
By default, \emph{index} and $(m,n)$ values are randomly matched, meaning one cannot infer that the explanations are similar for two \emph{index} values given that the \emph{index} values are similar. In our \emph{smooth} condition, we enforce a constraint in data generation so that the \emph{index} and $(m,n)$ values are ordered together, and similar \emph{index} values will have similar $(m,n)$ tuples (see Appendix~\ref{sec:app_experimental_details} for further details). 

We also measure the importance of including the \emph{index} in $x$, which is the easily computable feature linking query data points and explanations. Here, we use the default task setup, identical to that in RQ2, and we learn the retrieval model while using \textsc{TextCat}.

\paragraph{Results.} We show results across $n_\emph{task}$ and \emph{smoothness} in Fig.~\ref{fig:rq5}. The notable trend here is that learned retrieval clearly outperforms the baseline in the \emph{smooth} condition (by 7.6 points), while it only slightly outperforms the baseline in the \emph{non-smooth} condition (by 4.3 points). In terms of improvement over the fixed retriever, the differences are 8.1 points in the \emph{smooth} condition and 3.5 points in the \emph{non-smooth} condition. 
This result suggests that \textbf{learning to retrieve explanations will be particularly helpful when there is a sufficiently smooth notion of \emph{relevance} between data points and explanations}. The mechanism for this improvement is that, by retrieving explanations with similar \emph{index} values to the data point at hand, a model can guess the task parameters for the current data point since they will be close to the $(m,n)$ values in the retrieved explanations (fitting the definition of relevance in Sec.~\ref{sec:data_properties}).

Regarding the importance of the \emph{index}, we find that \textbf{for learning the retrieval to be possible, it is crucial that data and explanations are linked by an easily computable feature} such as the \emph{index}. Without including the \emph{index} in $x$, learned retrieval accuracy falls drastically from 98.6\% to 54.7\%.

\subsection{RQ6: Can explanations help models learn to use strong (causal, generalizable) features rather than weak ones?}
\label{sec:rq6}
\begin{figure}
\centering
 \includegraphics[width=.48\textwidth]{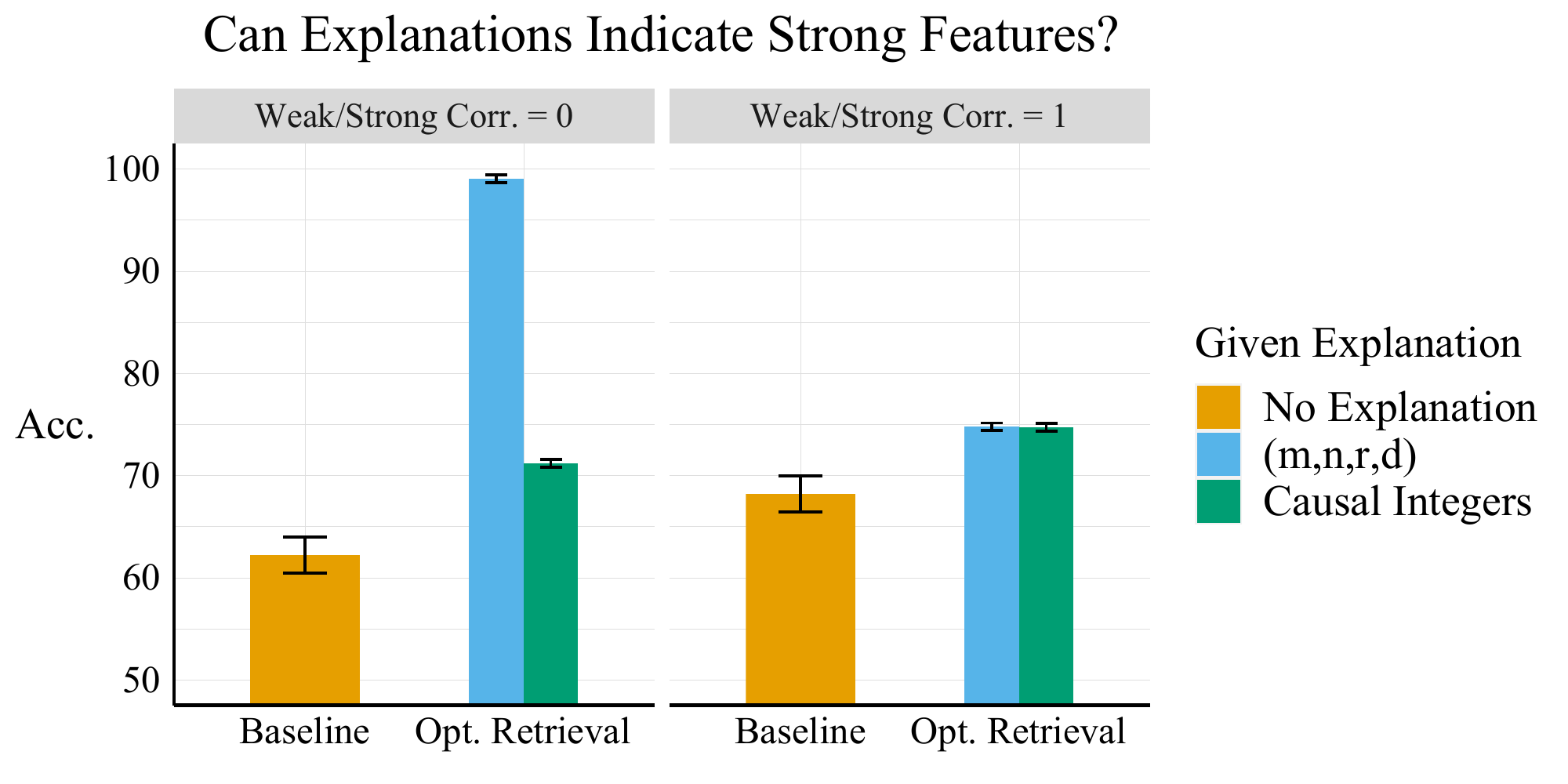}
 \vspace{-16pt}
\caption{(\textbf{RQ6}) Synthetic task accuracy grouped by the explanation kind and correlation between strong (causal) and weak (non-causal) features. In the Causal Integers condition, the model is always given the true pair of integers that must be counted.
}
\label{fig:rq6}
\vspace{-8pt}
\end{figure}

\paragraph{Design.} One especially intuitive use case for explanations is to help a model distinguish between strong, causal features and weak, spurious features. In this experiment, we vary the correlation between the strong and weak features in the training data along with the kinds of explanations that are retrieved by an optimal retrieval model. Recall that the strong feature in our task is $\mathbbm{1}[\mypound m{>}\mypound n]$ when $\emph{indicator}=1$ and $\mathbbm{1}[\mypound r{>}\mypound d]$ when $\emph{indicator}=2$, while the weak feature is drawn from the opposite integer pair's counts (refer to Sec.~\ref{sec:synthetic}). We emphasize that our strong and weak features are equally difficult to extract from the input; they differ only in that the strong feature causes the label, and the weak one does not. 
The explanations either match the familiar form, including all integers $(m,n,r,d)_{\emph{index}}$, or are restricted to include only the causal integers, $(m,n)$ if $\emph{indicator}{=}1$ and $(r,d)$ otherwise. When the strong-weak feature correlation is $1$, $m{>}n$ iff $r{>}d$ and $m{<}n$ iff $r{<}d$. When it is 0, the non-causal feature's relative count, i.e. $\mathbbm{1}[\mypound r{>}\mypound d]$ if $\emph{indicator}{=}1$, matches the strong feature's relative counts precisely half the time.

In all of these settings, the dev and test data are unaffected, meaning that a model with high test accuracy must have learned to use only the causal feature. We give additional results with the correlation varied between $0$ and $1$ in Fig.~\ref{fig:rq6b} in Appendix.~\ref{sec:app_experimental_details}.

\paragraph{Results.} We see in Fig.~\ref{fig:rq6} that, surprisingly, the only successful situation is when the original $(m,n,r,d)$ explanations are given and the strong-weak correlation is 0, under which the test accuracy is above 99\%. Note that, in the other settings, models most likely achieve around 75\% accuracy by predicting 1 when  $\mathbbm{1}[\mypound m{>}\mypound n] \vee \mathbbm{1}[\mypound r{>}\mypound d]$, since this strategy yields a test accuracy of 75\%.\footnote{Half of the time in the test data, the relative counts of $(m,n)$ and $(r,d)$ will agree by chance, meaning that predicting $\mathbbm{1}[\mypound m{>}\mypound n] \vee \mathbbm{1}[\mypound r{>}\mypound d]$ will yield 100\% accuracy. The other half of the time, the features will disagree, and this strategy yields 50\% accuracy. The overall test accuracy is then 75\%.} That is, \textbf{our ``causal feature" explanation fails to help when the strong and features are correlated and even when they are not}. This is surprising because we might expect that, when the correlation is 0, giving the causal feature should allow the model to succeed. After all, we may feel that we are effectively telling the model to count those two integers in every sequence. But \textbf{we risk anthropomorphizing the model whenever we suppose its interpretation matches our own}. From the model's standpoint, it sees a sequence of numbers \emph{whose relative counts are always unaffected by the two integers concatenated to the end of the sequence.} Our ``explanations" blend in seamlessly with the remainder of the sequence, except for the \texttt{[SEP]} token that happens to separate them. Hence we should not be so surprised that the model cannot use these explanations to pick out the causal feature; in fact, it may even be more impressive that the model \emph{does} succeed when the \emph{full-info} explanations are given. Evidently, the model learns a near-perfect interpretation of \emph{full-info} explanations with $5000$ training examples. 

\subsection{RQ7: How does the co-dependence between classifier and retrieval model influence the viability of joint training?}
\label{sec:rq7}
\begin{figure}
\centering
 \includegraphics[width=.48\textwidth]{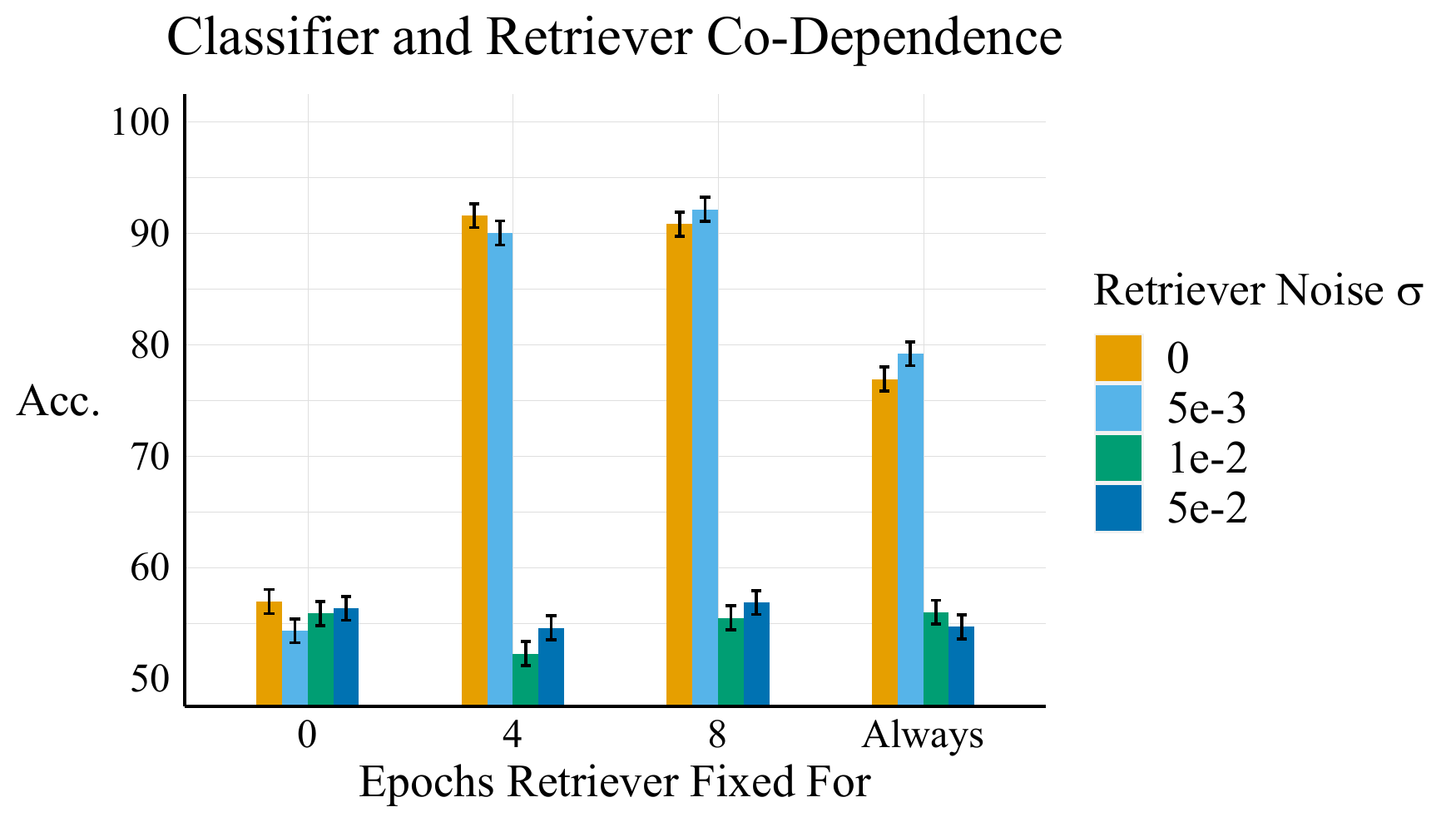}
 \vspace{-16pt}
\caption{(\textbf{RQ7}) The retrieval model must be fixed for some number of epochs for training to succeed. Meanwhile, degrading the quality of the initial retrieval by some amount of random noise can quickly render retrieval unlearnable.} 
\label{fig:rq7}
\vspace{-8pt}
\end{figure}

\paragraph{Design.} Since the learning signal for the retrieval model comes through the classifier, while the classifier relies on the retrieved explanations for its predictions, there is some co-dependence in their training dynamics. We further measure this co-dependence in a $4 \times 4$ design using evidential explanations with $\epsilon=2$. On one axis, we vary the number of training epochs for which only the classifier is trained and not the retrieval model, in values of $\{0, 4, 8, \infty \}$. On the other axis, we degrade the retrieval model by adding i.i.d. Normal noise to every parameter in the model, using $\sigma$ values of $\{0, \num{5e-3}, \num{1e-3}, \num{5e-2}\}$. To see the effect of choice of Sentence-BERT model, we perform another $3 \times 2$ experiment. Using either a randomly re-initialized RoBERTa-base, a standard pretrained RoBERTa-base, or the Sentence-RoBERTa-base model, we evaluate the performance with learned retrieval relative to fixed retrieval. 

\paragraph{Results.} We show results for the first experiment in Fig.~\ref{fig:rq7}. We find that the classifier must be warmed up, or, conversely, the retrieval model must be fixed, for some number of epochs to attain optimal performance. Jointly training both models from epoch $0$ on results in failed training runs. Meanwhile, adding noise to the initial retrieval model can quickly degrade its performance and render the retrieval unlearnable. Hence, \textbf{both retrieval model and classifier must reach some initial quality before training the other in order for joint training to succeed}.

As for the choice of pretrained retrieval model, we observe that retrieval is learnable only with Sentence-RoBERTa; retrieval is not learnable using RoBERTa-base, which performs about as poorly as using a randomly initialized retrieval model. These results are shown in Fig.~\ref{fig:rq7b} in Appendix~\ref{sec:app_experimental_details}. 

\subsection{RQ8: Does retrieval of explanations improve model performance on existing datasets?}
\label{sec:rq8}
\begin{table}[t]
    \centering
    \small
\begin{tabular}{l l c S[table-column-width=55pt, table-number-alignment=left]}
\toprule
Condition & Model & Acc. & {Effect Size}                    \\ 
\midrule
\addlinespace[4pt]
e-SNLI & & & \\
\addlinespace[2pt]
\hspace{5pt} $n{=}5000$ & RoBERTa & 84.83 (0.71)    &    \\
      & \textsc{TextCat} & 85.04 (0.71)    & .21 { (1.00)}         \\
\addlinespace[5pt]
\hspace{5pt} $n{=}$\num[group-separator={,}]{10000} & RoBERTa & 85.52 (0.70)    &    \\
      & \textsc{TextCat} & 86.03 (0.69)    & .51 { (0.98)}         \\
\addlinespace[5pt]
\hspace{5pt} $n{=}$\num[group-separator={,}]{50000} & RoBERTa & 87.90 (0.64)    &    \\
      & \textsc{TextCat} & 87.55 (0.65)    & -.35 { (0.92)}         \\
\addlinespace[5pt]
\hspace{5pt} $n{=}\emph{full}$ & RoBERTa & 91.06 (0.56)    &    \\
      & \textsc{TextCat} & 91.41 (0.55)    & .35 { (0.79)}         \\
\addlinespace[4pt]
SemEval & & & \\
\addlinespace[2pt]
\hspace{5pt} $n{=}5000$ & RoBERTa & 75.21 (1.62)    &    \\
      & \textsc{TextCat} & 75.73 (1.61)    & .52 { (2.29)}         \\
\addlinespace[5pt]
\hspace{5pt} $n{=}\emph{full}$ & RoBERTa & 76.94 (1.58)    &    \\
        & \textsc{TextCat} & 76.91 (1.59)    & -.03 { (2.24)}     \\
\addlinespace[4pt]
TACRED & & & \\
\addlinespace[2pt]
\hspace{5pt} $n{=}5000$ & RoBERTa & 84.24 (0.57)    &    \\
      & \textsc{TextCat} & 84.51 (0.57)    & .28 { (0.81)}         \\
\addlinespace[5pt]
\hspace{5pt} $n{=}$\num[group-separator={,}]{10000} & RoBERTa & 85.51 (0.55)    &    \\
      & \textsc{TextCat} & 86.14 (0.54)    & .63 { (0.78)}         \\
\addlinespace[5pt]
\hspace{5pt} $n{=}\emph{full}$ & RoBERTa & 88.29 (0.51)    &    \\
       & \textsc{TextCat} & 88.59 (0.50)    & .30  { (0.71)}      \\
\bottomrule
 \end{tabular}
 \vspace{-5pt}
\caption{Model accuracies for each dataset across training set sizes ($n$), with 95\% confidence intervals given in parentheses. We do not find retrieval of explanations to improve over baselines for any dataset and training set size.}
\label{tab:results}
\vspace{-9pt}
\end{table}

\paragraph{Design.} We test the retrieval-based model with three existing datasets: e-SNLI, TACRED, and SemEval. We also vary the training set size between values in \{5000, 10000, 50000\}, depending on the dataset, since the helpfulness of explanation retrieval could vary by the amount of available training data. Because \textsc{TextCat} achieves favorable results in our synthetic experiments, we use it as the conditioning mechanism here. Within each dataset, we tune $C$ and $k$ between values with the same product $Ck$, with the exception of e-SNLI using the full train set. For e-SNLI conditions with $n\leq 50000$, we select $(C{=}2,k{=}8)$. We use $(C{=}2,k{=}4)$ for e-SNLI with the full train set, given the expense of training retrieval in this setting. For most relation extraction settings we select $(C{=}2,k{=}4)$. See Appendix~\ref{sec:app_training} for further details.

Unlike in the synthetic data experiments, we consider adding $x_j$ and $y_j$ to the query data point along with retrieved explanation $e_j$, since explanations might best be interpreted in the context of the data they were given for. In tuning experiments we do not find any evidence for or against adding this extra information (see Table~\ref{tab:context_ablation} in Appendix~\ref{sec:app_training}). Here, we do add a textual representation of $y_j$ to the input $x_i$ along with retrieved explanations for relation extraction tasks, since these tasks have a higher number of classes. For e-SNLI, where $y_j$ can be easily inferred from the structure of explanations, we add only retrieved the explanations. 

Finally, for TACRED and SemEval, we compare to the ELV-M method in \citet{zhou2020towards}, which is \textsc{H-Mean} with ($C{=}10,k{=}1$) and fixed retrieval (discussed in Appendix~\ref{sec:app_experimental_details}). 

\paragraph{Results.} Shown in Table~\ref{tab:results}, \textbf{we see no statistically significant improvements from using explanation retrieval with any combination of dataset and training set size}. Across conditions, the effect sizes are slightly positive on average, but we are unable to assert any particular effect is positive. We also measure how accuracy varies across values of $k$ for finetuned models, but we do not find that increasing $k$ at test time improves accuracy (see Fig.~\ref{fig:rq8} in Appendix~\ref{sec:app_experimental_details}). In fact, the only statistically significant effect we see is from increasing the training set size. For example, doubling the TACRED training data from 5000 to 10000, increases the baseline accuracy by 1.28 ($p{=}.0017$).

Yet since we find that retrieval-based modeling succeeds in certain synthetic conditions, there must be a reason that the model fails to work well with datasets such as these. Using the results from this section, we speculate on the possible causes of this failure in Section \ref{sec:discussion} below.

\section{When Can Explanations Help?}
\label{sec:discussion}

In this section we take a position, based on our experimental findings, regarding the possible causes of the success of explanation retrieval in our synthetic task and its failure with e-SNLI, TACRED, and SemEval. 

Summarizing our experimental results, we suggest that \textbf{in principle, explanations can be helpful for modeling a task when:}
\begin{enumerate}[nosep, wide=0pt, leftmargin=*, after=\strut, label=(\arabic*)]
    \item The model can better infer relevant latent information given $x$ and the explanation, relative to using $x$ alone. Relevant latent information includes, for example, pertinent facts and task specifications that can assist with prediction. \emph{(RQ1, RQ2)} 
\end{enumerate}
However, this is not enough for explanations to be useful in practice. \textbf{Retrieval over explanations will be learnable to the extent that:} 
\begin{enumerate}[nosep, wide=0pt, leftmargin=*, after=\strut, label=(\arabic*), resume]
    \item Explanations are linked to query data points by an easily computable \emph{index} feature \emph{(RQ5)}, and
    \item There are explanations that are sufficiently \emph{relevant} across data points as to be useful for predicting labels for future data \emph{(RQ2, RQ5)}, and
    \item There is a known or identifiable \emph{interpretation} of the explanations by the classifier that yields a useful representation of the explanation \emph{(RQ3, RQ6)}, and
    \item Before training the retrieval model, the classifier reaches some sufficient quality \emph{(RQ7)}, and
    \item Before training the classifier, the initial retrieval model exhibits some sufficient quality \emph{(RQ7)}. 
\end{enumerate}
We wish to emphasize a few related results. One of the most intuitive use cases for explanations is to help a model distinguish between strong, generalizable features and weak, spurious features. But \textbf{explanations only help break ties between strong and weak features when the model already knows how to interpret the explanations}. When strong and weak features are perfectly correlated, we find that our synthetic explanations do not lead the model to select the causal feature more often than a non-causal one, even when using the optimal retrieval model. We only see that the model can learn to interpret the explanations when the features in question are not perfectly correlated.  
We suggest that, in the paradigm of large language model pretraining, this interpretation function will be meta-learned during pretraining. This behavior is clearly exemplified in GPT-3, which learns from pretraining to infer novel tasks from prompts that precede tasks in zero-shot settings \cite{gpt3}. Even GPT-2 learns some tasks such as summarization during pretraining, which can be elicited with the right prompt (e.g. ``tl;dr") \cite{radford_language_2019}. As we observe in our experiments, finetuning may allow the model to further identify the correct interpretation, provided that it is identifiable and sufficient training data is available.

It is also important to reiterate that \textbf{when using a retrieval model, the information that explanations provide can be inferred from the input alone}. A model need only learn the map between input and hidden information, rather than first using the input for retrieval and then interpreting the retrieved explanation. This is clearly possible in our synthetic task given the relationship between the \emph{index} and $(m,n,r,d)$ values.
The same situation will hold true of explanations for real-world tasks. Similar to our synthetic setting, if models can learn to retrieve explanations and then interpret them, they could instead learn to infer the latent information directly from the input. 
This property of tasks and explanations suggests that conditioning on explanations is a way to structure model computation, biasing it toward desirable functions, and away from difficult to learn functions. 
That is, as discussed in Sec.~\ref{sec:united_view}, we see explanations acting as \emph{priors} as well as simply inputs. 

So should we collect explanations to assist with solving tasks? At present, the answer is task-specific. In our synthetic task, it is far more helpful to have explanations for $5000$ training points than to have \num[group-separator={,}]{50000} unexplained points. On benchmark tasks such as e-SNLI, TACRED, and SemEval, we find that explanation retrieval does not yield statistically significant improvements in model accuracy, while using more unexplained data can lead to large improvements. We suggest that the reason for this lies somewhere in the six preconditions for explanation retrieval given above, and it will be useful in future work to develop a diagnostic procedure for further narrowing down the causes of model performance with and without explanations.

More broadly, we see two countervailing trends at work here. The first is that, as language models store more and more knowledge in their parameters, there will be less and less of a need for retrieved explanations to provide hidden information for tasks, though retrieval may still make ``accessing" this knowledge easier. In the other direction, we note that as language models become better at interpreting explanations and task descriptions, we will find that for some tasks performance is greatly boosted by having a good task description or set of explanations for example data points.

\section{Conclusion}

In this paper we present a formal framework for understanding the role of explanations in modeling, and we argue that explanations are most suitably used in a retrieval-based modeling approach, where past explanations are retrieved and used as model inputs for predicting future data points. We experimentally study the preconditions for explanations' usefulness in modeling, and based on results from our synthetic task, we suggest that the model must be able to better infer relevant latent information given the explanation and input, relative to using the input alone. For explanation retrieval to be learnable, we find that (1) explanations should be linked to query data points by an easily computable feature, (2) explanations should be relevant \emph{across} data points, (3) the interpretation of explanations by the classifier should be known or identifiable, and (4) the classifier and retrieval model must both be of some sufficient quality before the other is trained. When we test our method on three existing datasets (e-SNLI, TACRED, and SemEval), we find that explanations do not improve task performance, suggesting that these settings do not meet one of criteria outlined above. 

\section*{Acknowledgements}

We thank Miles Turpin and Ethan Perez for helpful discussion of the topics represented here, as well as Xiang Zhou and Prateek Yadav for feedback on this paper. This work was supported by NSF-CAREER Award
1846185, DARPA Machine-Commonsense (MCS) Grant N66001-19-2-4031, Royster Society PhD Fellowship, Microsoft Investigator Fellowship, and Google and AWS cloud compute awards. The views contained in this article are those of the authors and not of the funding agency.

\bibliography{main}
\bibliographystyle{icml2021}

\newpage
\appendix

\section{Training Details}
\label{sec:app_training}

\subsection{Data Preprocessing}

No preprocessing is applied to the synthetic data. For the three existing datasets, we use maximum sequence lengths as follows: For e-SNLI, we use a maximum sequence length of 120 tokens, with maximum lengths of 90 for $x$ and 60 for each explanation. For TACRED and SemEval, we use a max of 160 for the entire input, with a max of $80$ for $x$ and $60$ for $e$. We remove one explanation from the set of explained data points for TACRED after finding that it is given for a data point in the dev set.

We give additional examples of data points from each dataset in Table~\ref{tab:more_data_examples}.

\begin{table}[t]
\begin{center}
\small
\begin{tabular}{p{0.01\textwidth} p{0.4\textwidth}}
\toprule 
\addlinespace[4pt]
\mbox{\hspace{-5pt}e-SNLI Example 1} & \\
\addlinespace[1pt]
$x:$ & \emph{Premise:} After playing with her other toys, the baby decides that the guitar seems fun to play with as well. \emph{Hypothesis:} A blonde baby. \\
$y:$ & Neutral \\
$e:$ & Not all babies are blonde. \\
\addlinespace[4pt]
\mbox{\hspace{-5pt}e-SNLI Example 2} & \\
\addlinespace[2pt]
$x:$ & \emph{Premise:} A girl wearing a pink and black shirt and jeans fixes her hair before walking up the stairs. \emph{Hypothesis:} A girl has blonde hair. \\
$y:$ & Neutral \\
$e:$ & Not all girls have blonde hair. \\
\midrule
\addlinespace[4pt]
\mbox{\hspace{-5pt}SemEval Example 1} & \\
\addlinespace[1pt]
$x:$ & The SUBJ originates from an OBJ which transcends the speaker. \\
$y:$ & Entity-Origin \\
$e:$ & The phrase ''originates from an" occurs between SUBJ and OBJ and there are no more than four words between SUBJ and OBJ and OBJ follows SUBJ. \\
\addlinespace[4pt]
\mbox{\hspace{-5pt}SemEval Example 2} & \\
\addlinespace[2pt]
$x:$ & With one exception, the SUBJ emerged from the OBJ during hours of darkness. \\
$y:$ & Entity-Origin \\
$e:$ & The phrase ``emerged from the" occurs between SUBJ and OBJ and there are no more than four words between SUBJ and OBJ and SUBJ precedes OBJ. \\
\midrule
\addlinespace[4pt]
\mbox{\hspace{-5pt}TACRED Example 1} & \\
\addlinespace[1pt]
$x:$ & SUBJ's husband OBJ died in 1995. \\
$y:$ & Person-Spouse \\
$e:$ & Between SUBJ and OBJ the phrase ``'s husband" occurs and there are no more than five words between SUBJ and OBJ. \\
\addlinespace[4pt]
\mbox{\hspace{-5pt}TACRED Example 2} & \\
\addlinespace[2pt]
$x:$ & SUBJ is married to OBJ and is the father of three sons. \\
$y:$ & Person-Spouse \\ 
$e:$ & There are no more than four words between SUBJ and OBJ and the phrase ``is married to" appears between SUBJ and OBJ. \\
\bottomrule
\end{tabular}
\end{center}
\vspace{-5pt}
\caption{Additional example data points from three existing datasets.}
\vspace{-7pt}
\label{tab:more_data_examples}
\end{table}

\subsection{Runtimes.}

Regarding training times, we run most experiments on a single NVIDIA RTX 2080 GPU, with runtimes as follows: 4.0 hours for 40 epochs of the no-retrieval RoBERTa-base using the synthetic dataset; 5.7 hours for 40 epochs of RoBERTa-large in the same setting; 8.6 hours for 20 epochs of learned retrieval with RoBERTa-base models on synthetic data; 32.9 hours for 10 epochs of learned retrieval with TACRED. Several adjacent experimental conditions can be easily extrapolated here given the training sizes for these conditions. Lastly, we run our full-data e-SNLI condition with learned retrieval for 5 epochs on a single Tesla P100 GPU, which takes 7 days to run. 

\subsection{Training Hyperparameters and Analysis}

\begin{figure}
\centering
 \includegraphics[width=.48\textwidth]{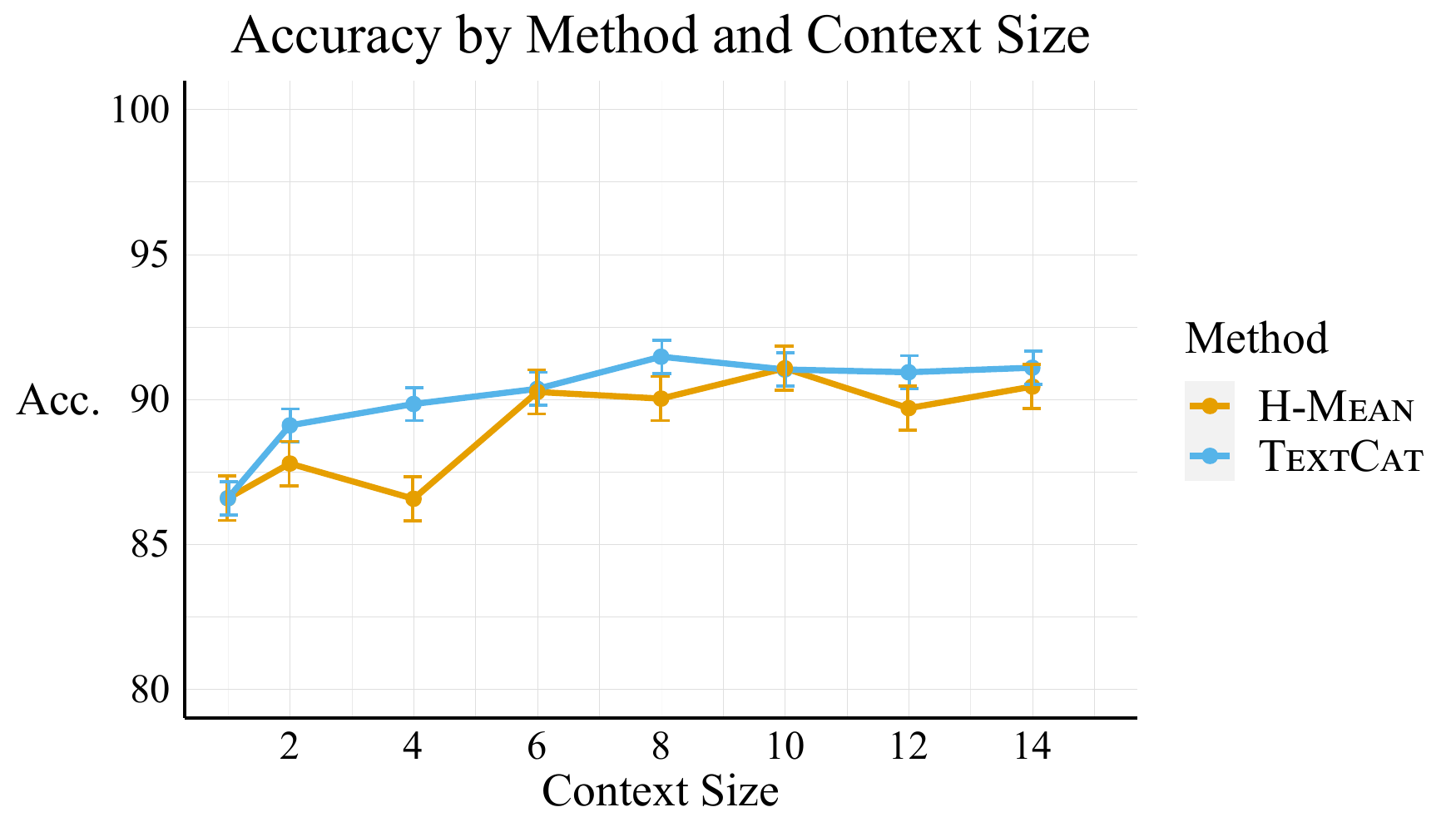}
 \vspace{-16pt}
\caption{(\emph{Training Hyperparameters and Analysis.}) Learned retrieval accuracy by $C$ and conditioning mechanism, using $k=1$ and optimal retrieval of \emph{evidential} explanations.} 
\label{fig:by_c}
\vspace{-8pt}
\end{figure}

\begin{figure}
\centering
 \includegraphics[width=.48\textwidth]{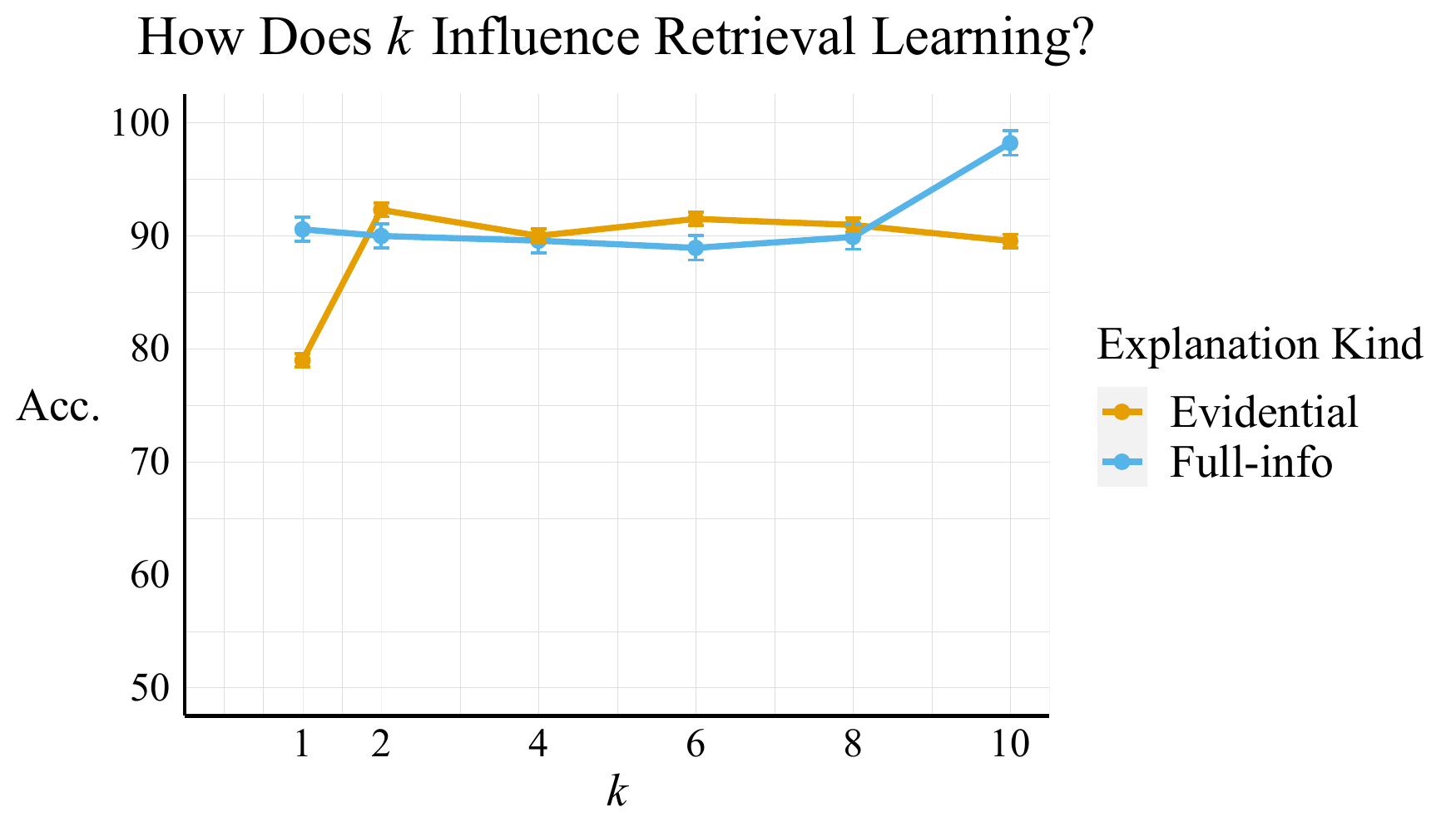}
 \vspace{-16pt}
\caption{(\emph{Training Hyperparameters and Analysis.}) Learned retrieval accuracy by $k$, with $C=1$, for \emph{full-info} and \emph{evidential} explanations.} 
\label{fig:by_k}
\vspace{-8pt}
\end{figure}

For optimization, we use AdamW with a learning rate of \num{1e-5} and gradient norm clipping at norm 1. For the LR, we use a linear warmup and decay schedule peaking at 10\% of the training steps for experiments with synthetic data and at 1\% for experiments with existing datasets (given the larger training set sizes). The batch size is set to 10 across all experiments. 

We decide how often to rebuild the representations of training explanations while learning the retrieval model by tuning across frequency values in the range \{10\%, 20\%, 33\%, 50\%, 100\%\} (i.e. to rebuild at this percentage of every epoch), as well as never rebuilding. In our synthetic setting, the only noticeable drop in performance comes from never rebuilding. As long as representations are re-encoded at least as often as every epoch, we notice no difference in final test accuracy, though in early experiments we observed that rebuilding more often improved training stability. To err on the safe side of training stability, we re-encode the representations every 20\% of each epoch in all experiments except e-SNLI with full data, where we re-encode every 30\% of each epoch. 

Additionally, we use the stop-gradient function when computing the gradient of $p_\eta(e|x)$ as follows:
\begin{align*}
    \nabla_\eta \exp{(\textrm{sg}[f_\eta(e)]^Tf_\eta(x))},
\end{align*}
meaning that we do not differentiate through the explanation embeddings, but only through the query data point embeddings. In early experiments, we found that this decision contributed to training stability, while improving computational efficiency, and we confirm that we observe no differences in model accuracy as a result. 

We measure the relationship between the context size $C$ and performance on evidential explanations, using the optimal retrieval model and comparing between conditioning mechanisms. The results are shown in Fig.~\ref{fig:by_c}. We see that, with each method, a larger value of $C$ is preferable up to around $8$ or so, after which performance plateaus. 

Regarding the value of $k$, we see in Fig.~\ref{fig:by_k} that training performance can be sensitive to the chosen value for this hyperparameter. It appears that one should try to select as high a value of $k$ as possible, all else equal. Though since this parameter increases the number of forward passes during training by a factor of $k$, there is a trade-off between the available compute budget and the value of $k$ in practice. 

\subsection{Model Selection in Experiments}

\begin{table}[t]
    \centering
    \small
\begin{tabular}{l l c}
\toprule
Condition & Model & Acc.                    \\ 
\midrule
\addlinespace[4pt]
e-SNLI & & \\
\addlinespace[2pt]
\hspace{5pt} $n{=}$\num[group-separator={,}]{10987} & \textsc{TextCat-E} & 87.17 (0.66)     \\
      & \textsc{TextCat-YXE} & 87.11 (0.66) \\
\addlinespace[4pt]
SemEval & & \\
\addlinespace[2pt]
\hspace{5pt} $n{=}$\num[group-separator={,}]{7016} & \textsc{TextCat-YE} & 75.25 (2.99)      \\
      & \textsc{TextCat-YXE} & 75.75  (2.97)   \\
\addlinespace[4pt]
TACRED & & \\
\addlinespace[2pt]
\hspace{5pt} $n{=}$\num[group-separator={,}]{68124} & \textsc{TextCat-YE} & 87.49 (0.43)  \\
& \textsc{TextCat-YXE} & 87.31 (0.43) \\  

\bottomrule
 \end{tabular}
 \vspace{-5pt}
\caption{Ablation across the retrieved variables: the suffix to \textsc{TextCat} indicates which retrieved variables are included in the model input. We do not find that including $x$ improves model performance, so we use only $e$ or $(y,e)$, depending on the task.}
\label{tab:context_ablation}
\vspace{-9pt}
\end{table}

In general, within a single training run, we select the model that achieves the best dev set accuracy as measured at the end of each training epoch.

With our synthetic task, we observe some training instability in a few conditions, particularly in those where we are degrading the training model (RQ7). On such occasions, training fails after a few epochs and model accuracy trends toward 50\% (random performance).  These occurrences are easily noticeable, so we rerun these experiments with a different seed in order to report results from a stable run, and typically we find that stable training dynamics can be obtained from just one other seed.

For the existing datasets, we run three model seeds for the baseline and each of the hyperparameter conditions, except for when using at least \num[group-separator={,}]{50000} training data points, where we run only one seed. We also use only one seed when ablating across which retrieved variables to include in the model input (i.e. whether to include the retrieved $x$ in the model input). To select one model from three seeds for a given condition, we pick based on the highest dev performance.

The results for ablating across the retrieved variables to include as model inputs are shown in Table~\ref{tab:context_ablation}. Note that we test the effect of adding $x$ to $e$ for e-SNLI and the effect of adding $x$ to $(y,e)$ for relation extraction tasks, since $y$ is easily inferred from $e$ for e-SNLI \cite{hase2020leakage}. In these experiments, we roughly control for the sequence length, meaning that for relation extraction tasks, we use $C=1$ when $x$ is present and $C=2$ when it is not, while for NLI we use $C=5$ without $x$ and $C=2$ with $x$. These experiments all use $k=4$. We do not find any statistically significant differences in dev set accuracy across any of the conditions. Hence, we proceed with using $(y,e)$ for relation extraction tasks and $e$ for NLI.

When tuning over $C$ and $k$ with the existing datasets, we use the following values for each condition: For relation extraction tasks, we tune over values in \{($C{=}2,k{=}4$),($C{=}1,k{=}8$)\}, and for NLI we tune over  \{($C{=}4,k{=}4$),($C{=}2,k{=}8$)\} when $n<50000$. We tune separately for each training set size configuration. We select the hyperparameters to use based off of the best dev set accuracy achieved from three seeds in each condition. We report results from a single run of $(C{=}2,k{=}4)$ for e-SNLI with the full training data.

\section{Experimental Details And Additional Results}
\label{sec:app_experimental_details}

In this section, we describe experimental details and hyperparameters for particular experiments, organized by research question, as well as additional results accompanying some research questions. Lastly, we discuss hypothesis testing procedures. Unless otherwise stated, additional experiments below use the default synthetic task parameters, given in the Synthetic Task section in the main paper. 

\subsection{RQ1: When can models solve our synthetic problem by inferring each sequence's task, and when must they be given the task information?}
\begin{figure}
\centering
 \includegraphics[width=.48\textwidth]{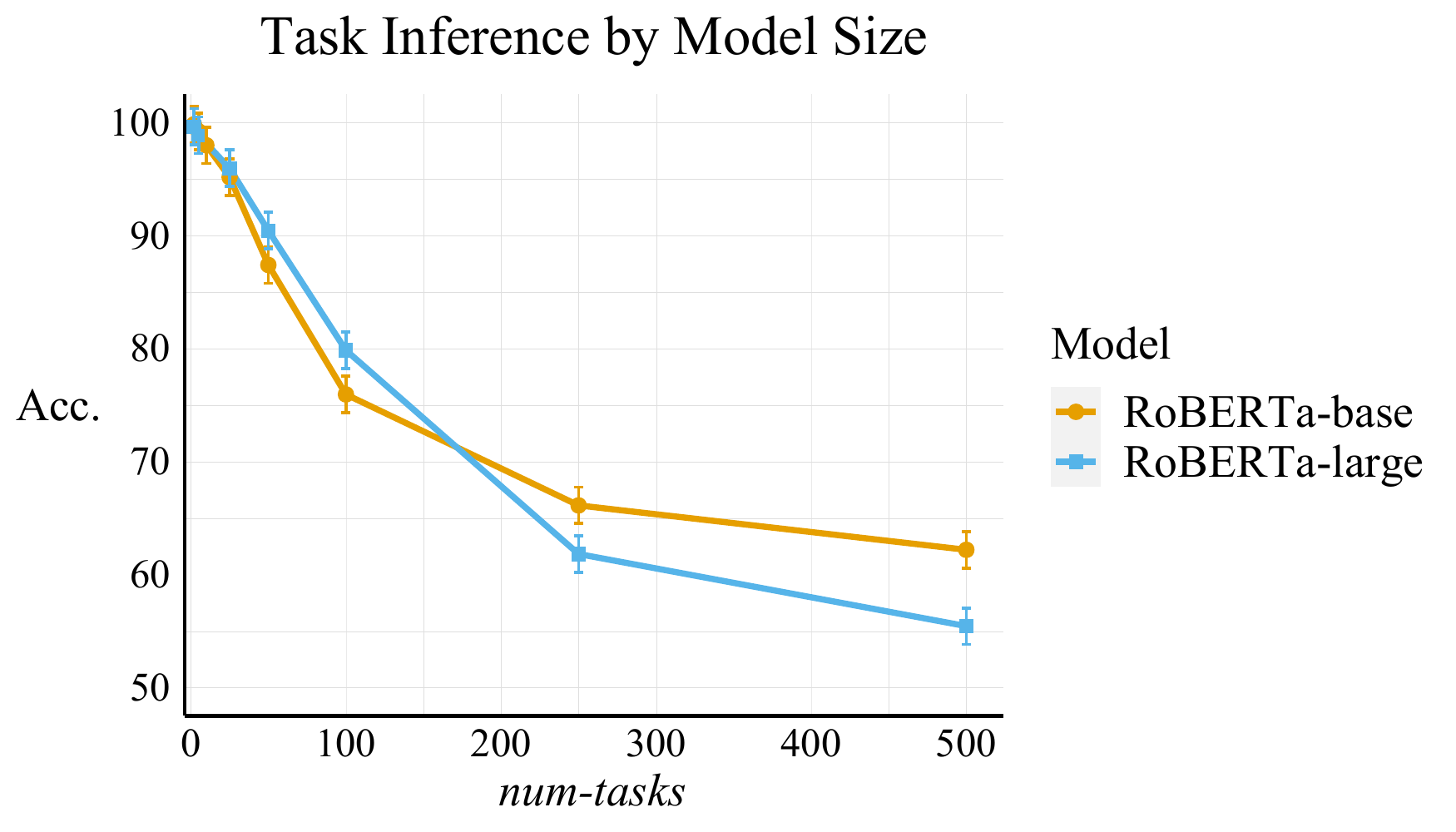}
 \vspace{-16pt}
\caption{(\textbf{RQ1}) Synthetic task accuracy as a function of \textit{num-tasks}, by model size.} 
\label{fig:rq1b}
\vspace{-8pt}
\end{figure}

\begin{figure}
\centering
 \includegraphics[width=.48\textwidth]{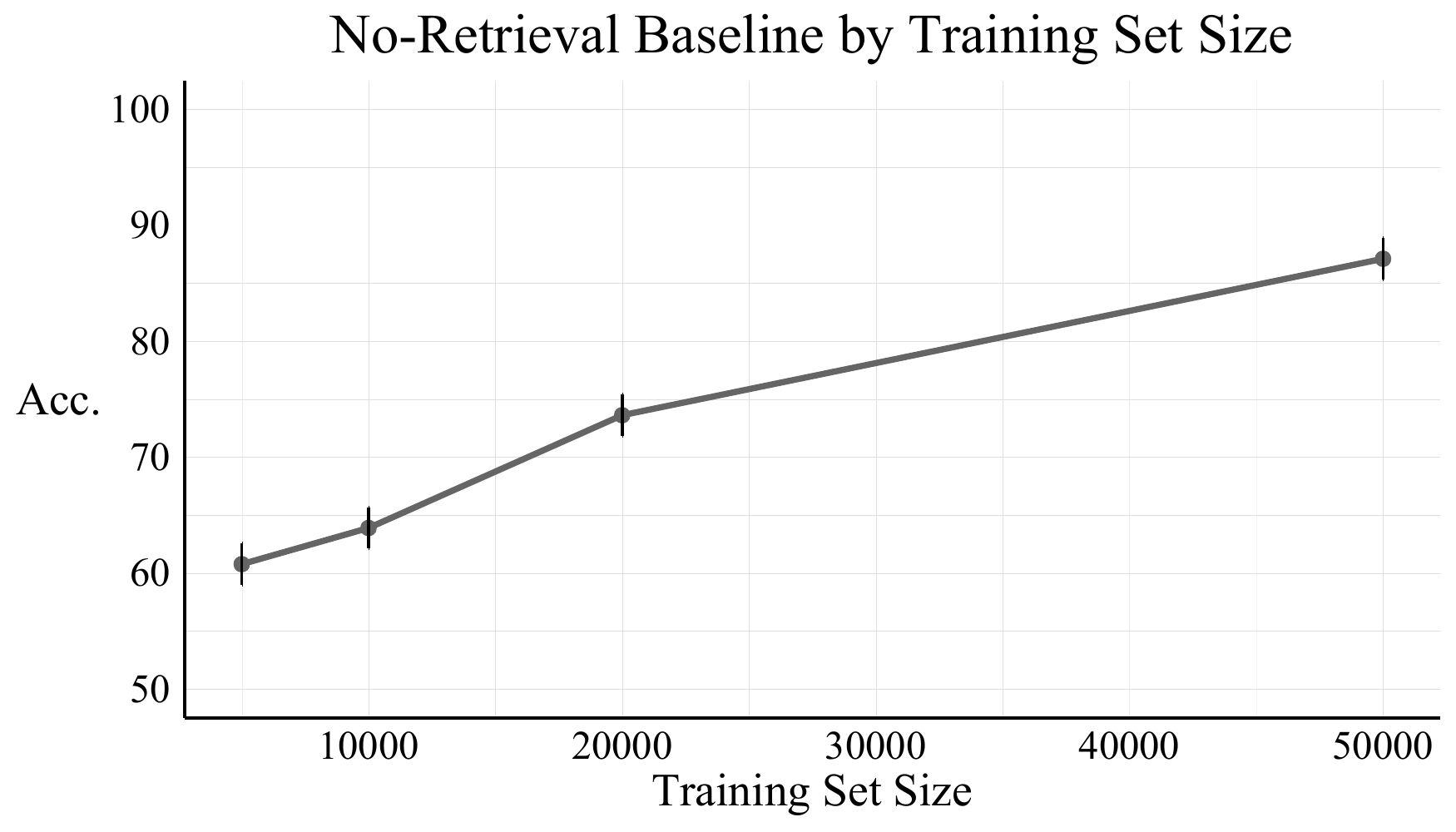}
 \vspace{-16pt}
\caption{(\textbf{RQ1}) Synthetic task accuracy as a function of the training set size, without explanation retrieval.} 
\label{fig:rq1c}
\vspace{-8pt}
\end{figure}

We show results for additional model choices and sizes of the training datasets here. In Fig.~\ref{fig:rq1b}, we see that RoBERTa-large outperforms RoBERTa-base only when the number of tasks in the training data is relatively small. After \emph{num-tasks}${\geq}100$, RoBERTa-base is the better choice. Here, we train models for 40 epochs with a LR of \num{1e-5}, though for \emph{num-tasks}$\in\{250,500\}$ with RoBERTa-large, we have to train for 60 epochs with a LR of \num{1e-6} in order for training to converge. 

In Fig.~\ref{fig:rq1c}, we see that performance scales well with the available training data for our synthetic task. RoBERTa-base reaches an accuracy of $87.11$ with \num[group-separator={,}]{50000} training points.

\subsection{RQ2: Can retrieval of past explanations enable a model to solve our task?}

In these experiments, we always freeze the retriever for the first two epochs of training and train for a total of 20 epochs.

\subsection{RQ3: Can models aggregate information across explanations for better prediction?}

Here, we freeze the retriever for the first five epochs of training and train for a total of 25 epochs.

\subsection{RQ4: What is the best way to compute explanation representations for prediction?}
\begin{figure}
\centering
 \includegraphics[width=.48\textwidth]{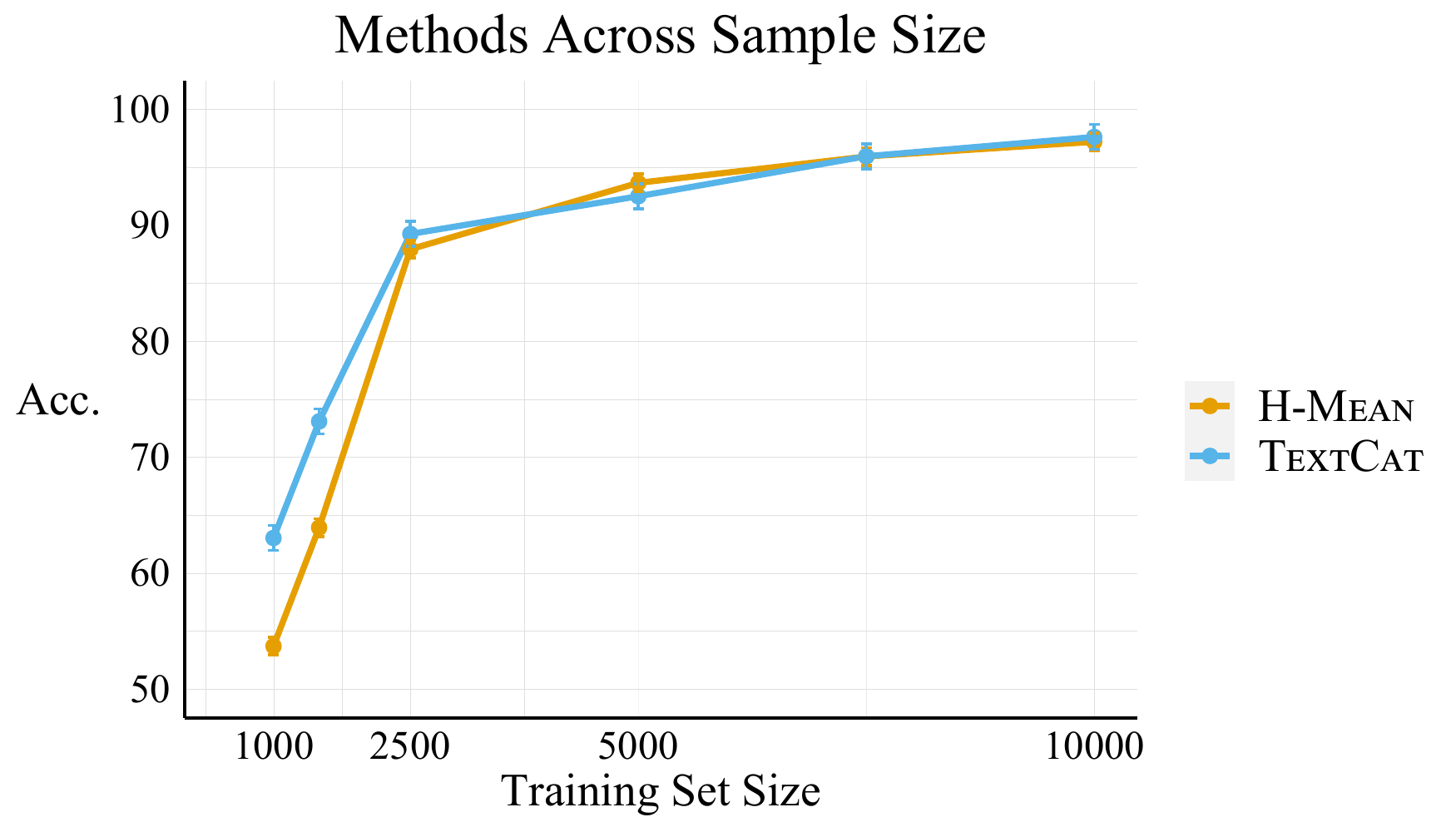}
 \vspace{-16pt}
\caption{(\textbf{RQ4}) Synthetic task accuracy across training set sizes and method, using optimal retrieval.} 
\label{fig:rq4}
\vspace{-6pt}
\end{figure}

In Fig.~\ref{fig:rq4}, we see that \textsc{TextCat} outperforms \textsc{H-Mean} at smaller training set sizes. \textsc{TextCat} achieves higher accuracy than \textsc{H-Mean} by 9.3 points for $n=1000$ and 9.2 points for $n=1500$, though the gap shrinks to 1.3 points at $n=2500$ and the methods perform equally well after $n=5000$. In these experiments we use the optimal retrieval model. 

\subsection{RQ5: What makes an explanation relevant across data points? What enables a retrieval model to find relevant explanations for a new data point?}

In these experiments, we use $C=1$ and $k=12$, and we make changes to the default data properties regarding the $n_\emph{task}$ and \emph{smoothness}. In order to achieve a smooth function from \emph{index} to $(m,n)$, we first order the domain and codomain and then match them up one-to-one. To order $(m,n)$ tuples, we sort by $m$ first and then $n$. The result is that when two explanations have similar \emph{index} values, their $m$ values are very likely to be close together, and their $n$ values will probably be close together. Note that in this experiment, we sample $(m,n)$ not from $unif([1,100]^2)$, but rather we draw the valid $(m,n)$ tuples in increasing order starting from the first valid tuple, $(1,2)$. We use the same $(m,n)$ sampling scheme for the baselines and the \emph{non-smooth} condition. The only difference in the \emph{non-smooth} conditions is the lack of ordering in the domain and codomain before they are matched up. This allows for more precise inference as to the task parameters when retrieving an explanation with a similar \emph{index} to a data point at hand. 

\subsection{RQ6: Can explanations help models learn to use strong features rather than weak ones?}
\begin{figure}[t]
\centering
 \includegraphics[width=.48\textwidth]{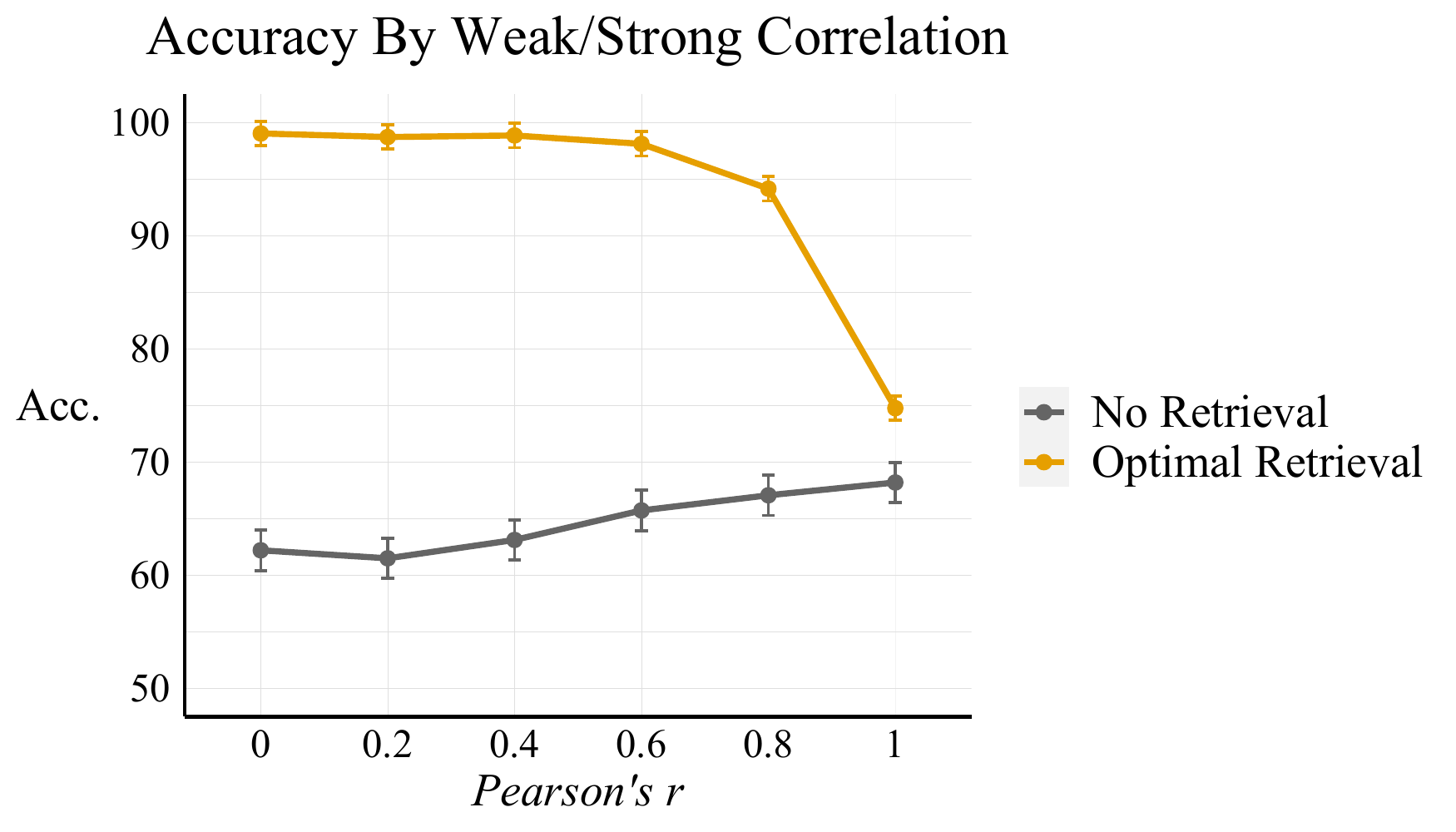}
 \vspace{-16pt}
\caption{(\textbf{RQ6}) Synthetic task accuracy across correlation levels between the strong and weak feature, with and without optimal retrieval.} 
\label{fig:rq6b}
\vspace{-8pt}
\end{figure}

\begin{figure}[t]
\centering
 \includegraphics[width=.48\textwidth]{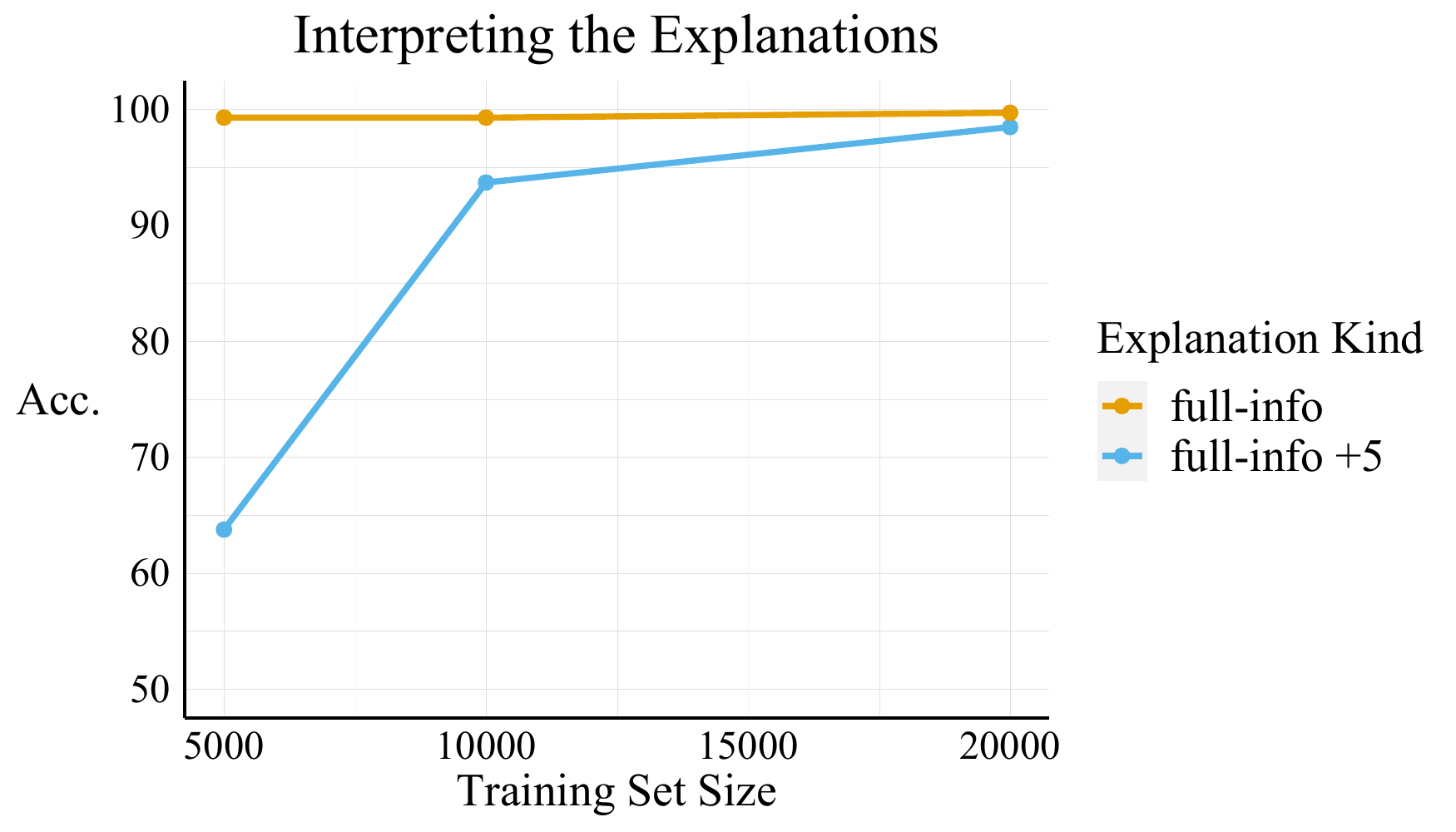}
 \vspace{-16pt}
\caption{(\textbf{RQ6}) Even a simple function on explanations can render them difficult for the model to interpret, though the correct interpretation is identified with more data.} 
\label{fig:rq6c}
\vspace{-8pt}
\end{figure}

We give additional results with the strong-weak feature correlation varied between $0$ and $1$ in Fig.~\ref{fig:rq6b}, using the training hyperparameters for RQ2. Using the \emph{full-info} explanation with optimal retrieval, we see the model continues to perform well as long as the features are not perfectly correlated. Interestingly, the no-retrieval condition's performance rises as the correlation increases, though it never matches the retrieval condition's performance. Since the performance of optimal retrieval is above the baseline but not greater than 75\% when the features are perfectly correlated, the explanations are not helping decide whether to use the strong or weak feature, but they are helping these features be used in the first place (see the footnote in the results for RQ5 in the main body).

It may even be surprising that the \emph{full-info} explanations are useful when the strong-weak correlation is 0, since the Causal Integer explanations are not. In Fig.~\ref{fig:rq6c}, we see that, while the correlation is 0, some explanations may be hard to interpret when a small amount of training data is available, but as more data is available the correct interpretation is identified. Here, we simply add 5 to each integer in the \emph{full-info} explanations. Using optimal retrieval, models struggle to correctly interpret these explanations at a low sample size, but with more data the correct interpretation is identified.

\subsection{RQ7: How does the co-dependence between classifier and retrieval model influence the viability of joint training?}
\begin{figure}
\centering
 \includegraphics[width=.48\textwidth]{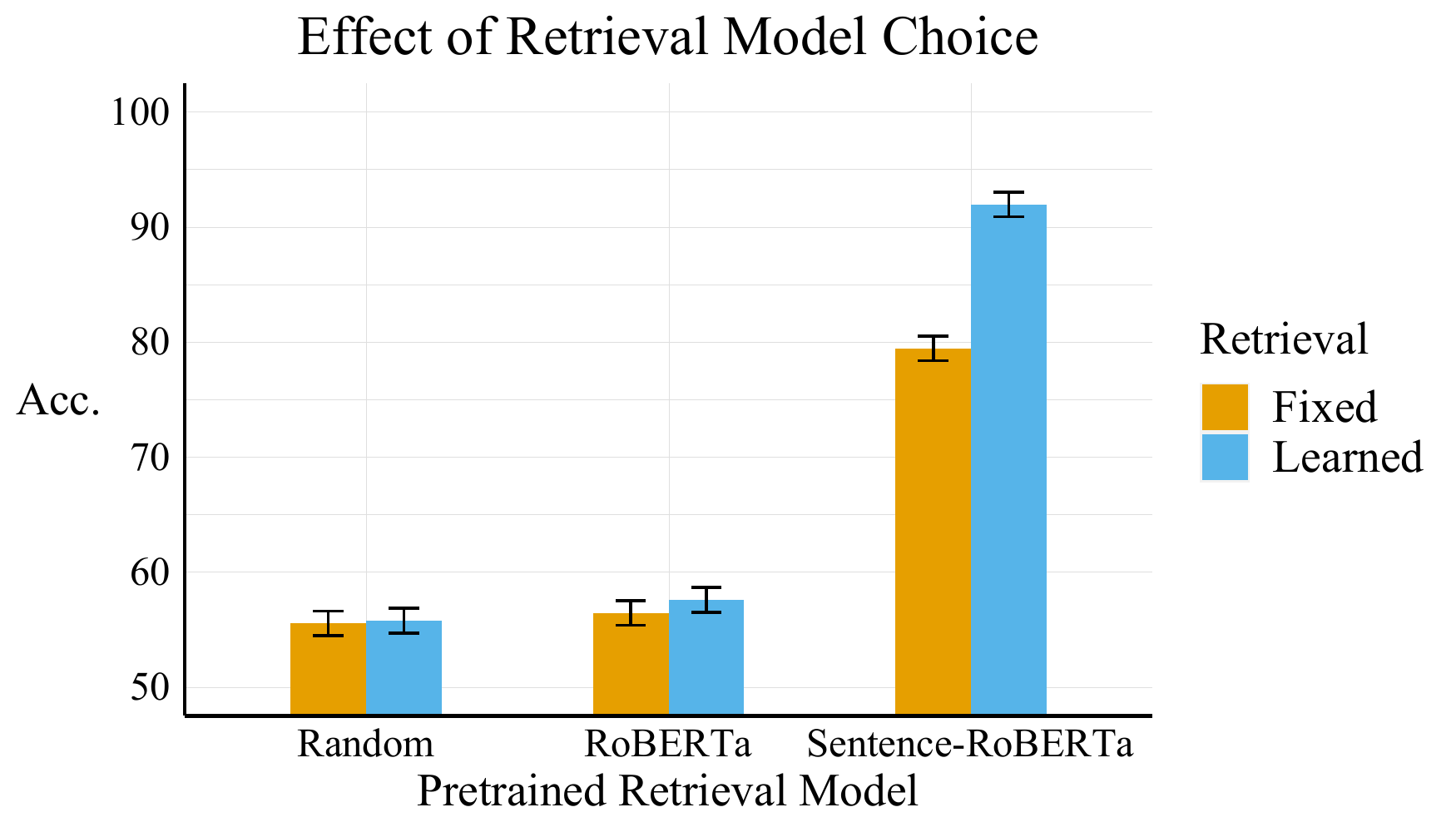}
 \vspace{-16pt}
\caption{(\textbf{RQ7}) Model performance by choice of retriever, with \emph{evidential} explanations. Using a pretrained Sentence-BERT model is vital to the success of learning a retrieval model in our synthetic task.} 
\label{fig:rq7b}
\vspace{-8pt}
\end{figure}

Hyperparameters in experiments for this RQ match those for RQ3. In Fig.~\ref{fig:rq7b}, we show the effect of the retrieval model choice on the viability of learning retrieval. As in the main body, we also use evidential explanations with $\epsilon=2$. We find that it is necessary to use a pretrained Sentence-RoBERTa model. Simply using a pretrained RoBERTa-base model will not suffice for learning retrieval with our synthetic task. Surprisingly, this condition cannot outperform even a randomly initialized model with an identical architecture. This could be due to the fact that we use the mean token pooling and cosine similarity that the Sentence-BERT models were trained with.

\subsection{RQ8: Does retrieval of explanations improve model performance on existing datasets?}
\begin{figure}
\centering
 \includegraphics[width=.48\textwidth]{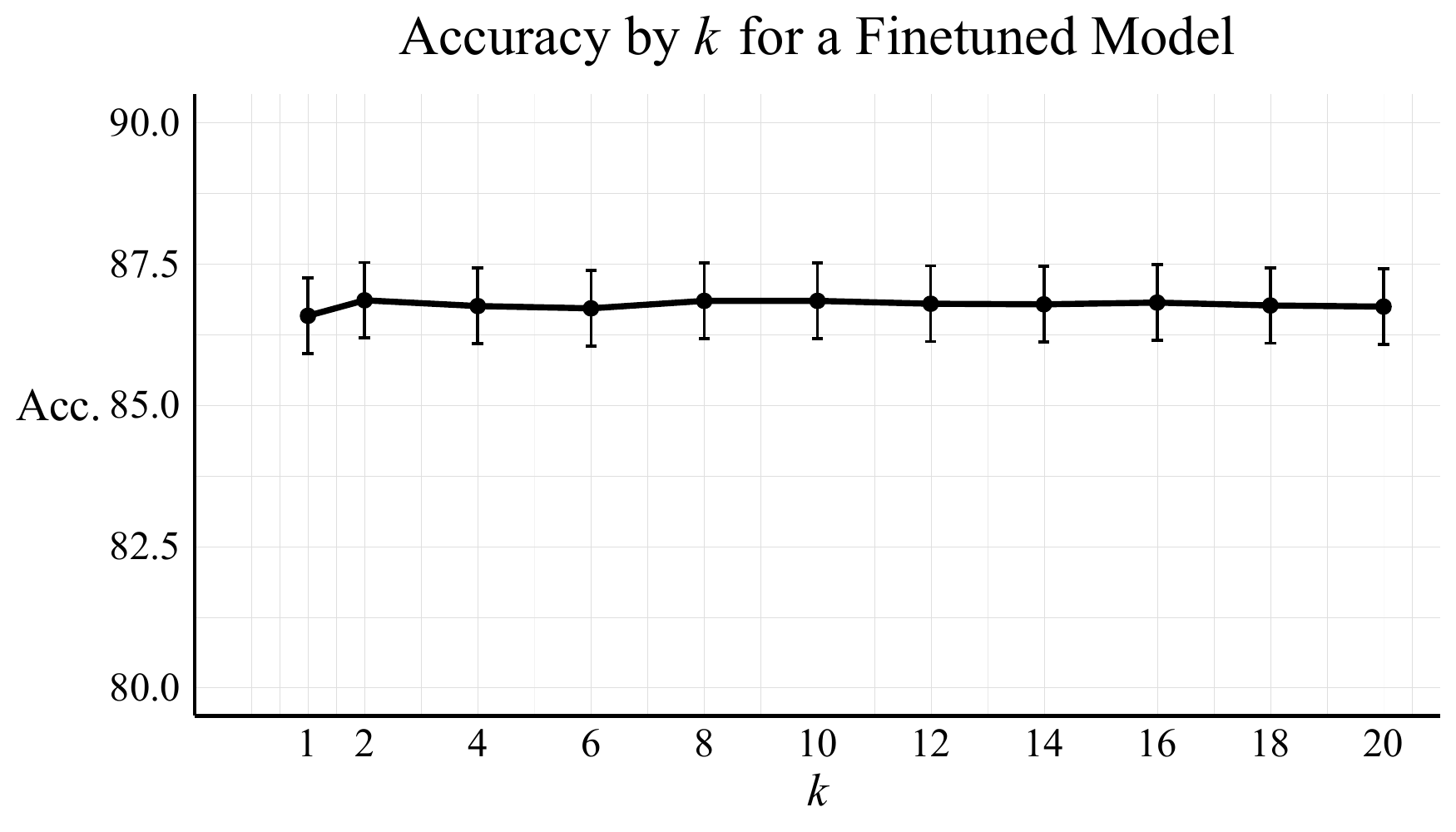}
 \vspace{-16pt}
\caption{(\textbf{RQ8}) Dev set accuracy across $k$ for the retrieval model on e-SNLI using $10000$ training points.} 
\label{fig:rq8}
\vspace{-8pt}
\end{figure}

We train for: 5 epochs when using the full e-SNLI training set; 20 epochs when using $n{\leq}10000$ for any dataset; and 10 epochs for other larger values of $n$. 

In Fig.~\ref{fig:rq8}, we show the result of varying the value of $k$ used to calculate dev set accuracy for the retrieval model in the e-SNLI with $n=10000$ condition. We see no meaningful changes in dev set accuracy across values of $k$ from 1 to 20, showing that increasing $k$ at test time is not a reliable way to improve retrieval model accuracy in this setting.

Lastly, we observe that the ELV-M condition from \citet{zhou2020towards}, which is \textsc{H-Mean} with fixed retrieval and $(C{=}10,k{=}1)$, does not outperform baselines on TACRED and SemEval. The approach obtains 87.99\% on TACRED, where our baseline is 88.29\%, and 76.46\% on SemEval, where the baseline is 76.94\%. Besides using RoBERTa models instead of BERT, one change we make from the implementation in \citet{zhou2020towards} is to disallow for data points' own explanations to be conditioned on when predicting their labels, although this is not relevant for predicting test points in either dataset.

\subsection{Confidence Intervals and Hypothesis Testing}
\begin{figure}
\centering
 \includegraphics[width=.48\textwidth]{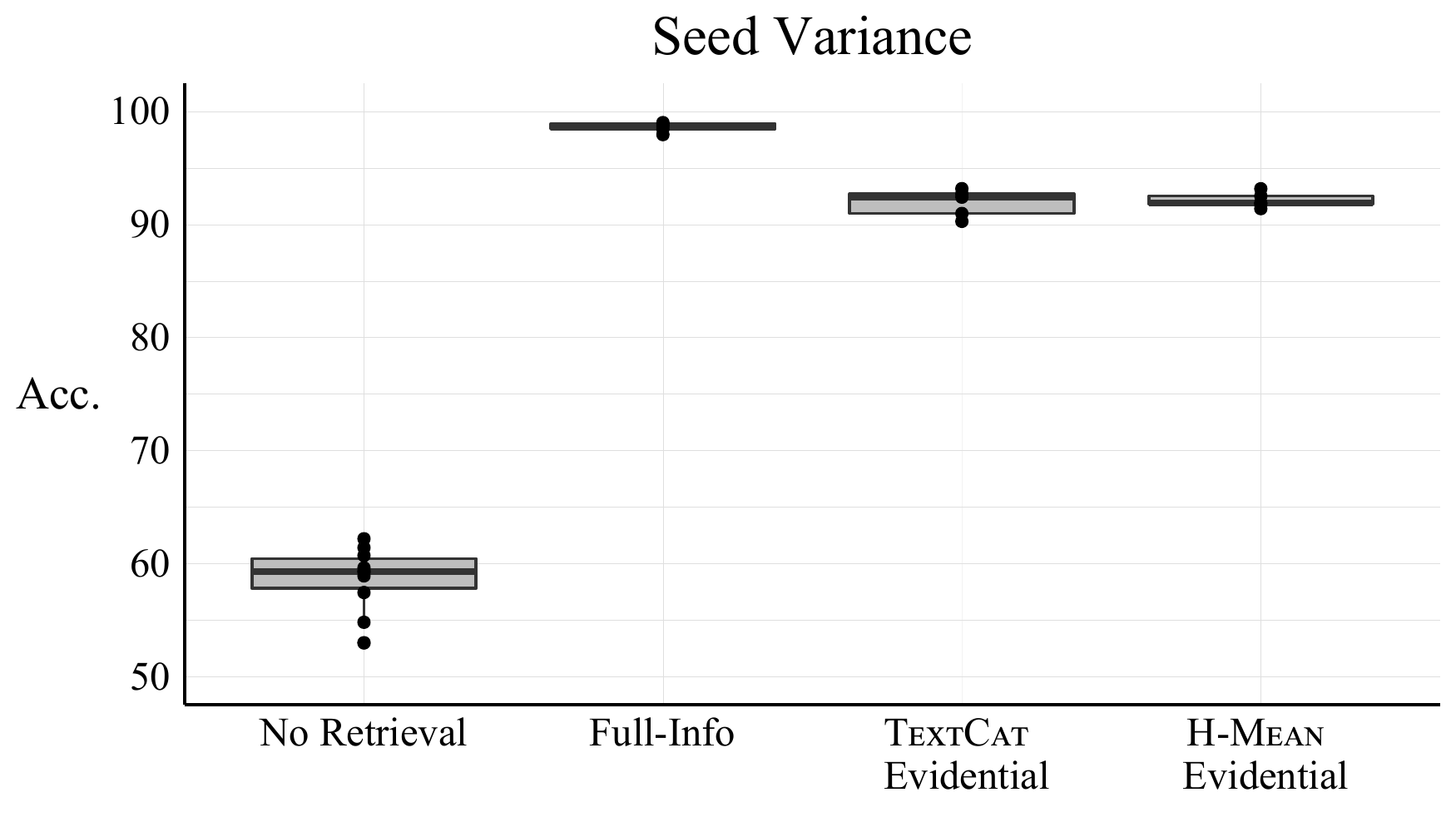}
 \vspace{-16pt}
\caption{Seed variance for some representative experimental conditions.} 
\label{fig:seed_var}
\vspace{-8pt}
\end{figure}

We compute confidence intervals for our synthetic data tasks to represent \emph{seed variance} around some mean seed performance, while confidence intervals and associated hypothesis tests for existing datasets represent \emph{sample variance.} With synthetic data we represent seed variance in figures rather than sample variance because the sample variance is fairly low with 
\num[group-separator={,}]{50000} test points and could be driven arbitrarily low with more generated test points. For instance, the 95\% confidence interval for a model accuracy of 90\% would be $\pm 0.26$. 

To calculate seed variance, we run 10 random seeds for our baseline condition (no-retrieval) with the default synthetic task setup. Then we run 5 runs with learned retrieval using (1) \textsc{TextCat} with \emph{full-info} explanations, (2) \textsc{TextCat} with \emph{evidential} explanations, and (3) \textsc{H-Mean} with \emph{evidential} explanations. The results of these runs are shown in Fig.~\ref{fig:seed_var}. We then assume that seed variance is invariant across experimental factors not related to the choice of conditioning method or explanation and assign 95\% confidence intervals across experimental conditions based on these four representative conditions. We prioritize assignments based on the explanation kind (\emph{full-info} vs. \emph{evidential} or \emph{recomposable}), then by conditioning mechanism, when for instance some conditions use combinations of methods and explanation kinds not represented in these conditions. We assume these invariances in order to efficiently calculate seed variance. Running 5 seeds per retrieval condition and 10 per non-retrieval would increase the number of synthetic data experiments in this paper from 172 to 1035. In synthetic data experiments, we comment on effects far larger than the confidence intervals and do not conduct hypothesis tests. 

The confidence intervals shown for model accuracies on existing datasets are 95\% confidence intervals on the underlying binomial probability. The hypothesis tests conducted for RQ8 are two-sided difference in binomial means tests. 

\section{Synthetic Task Generative Process}
\label{sec:synthetic_task_generation}

The required parameters to the data generation include: (1) a training sample size \emph{sample-size} and (2) \emph{num-tasks}, the number of unique integer pairs to be counted, or, equivalently, the number of points per \emph{index}, $n_\emph{task}$. In all experiments, we use a maximum integer value of 100 to appear in the sequences, and a maximum \emph{index} value of \num[group-separator={,}]{10000}. We give the general generative process below. Note that the dev and test sets are constructed with the extra constraint that sequences must not appear in the training data. Further note that this is the generic version of generative process, and in some experiments the process is altered. For example, in RQ5, \emph{indicator} is always 1 and the construction of the map from \emph{index} values to $(m,n)$ tuples occurs in a special way described in the experimental design for RQ5.

\begin{enumerate}[itemsep=3pt, wide=0pt, leftmargin=*, after=\strut]
    \item Sample $\{index_t\}_{\tau=1}^{\emph{num-tasks}}$ from the uniform distribution over integers \{1,...,10000\} without replacement.
    \item Sample $\{(m,n,r,d)_{t}\}_{\tau=1}^{num\textrm{-}tasks}$ from the uniform distribution over integers, $unif([1,100]^4)$, without replacement and requiring that $m\neq n \neq r \neq d$. 
    \item Define the set $\{(\emph{index}, m, n,r,d)_\emph{index})\}$ for \emph{index} and $(m,n,r,d)$ drawn from their respective sets, without replacement, in an arbitrary order. 
    \item Compute the number of points per \textit{index}, $n_\emph{task}~{=}~\emph{sample-size} \ // \ \emph{num-tasks}$.
    \item For each $\emph{index} \in \{\emph{index}_t\}_{\tau=1}^{\emph{num-tasks}}$:
    \begin{enumerate}[nosep, wide=0pt, leftmargin=*, after=\strut]
        \item 
        Sample a vector of length $n_\emph{task}$, balanced between $1$s and $2$s, that gives the values of $\{\emph{indicator}_p\}_{p=1}^{P}$ for the $P$ points with that \textit{index}. 
        \item Sample a vector of length $n_\emph{task}$, balanced between 0s and 1s, representing whether the features $\mathbbm{1}[\mypound m{>}\mypound n]$ and $\mathbbm{1}[\mypound r{>}\mypound d]$ should correlate  (1 implies they are equal, and 0 unequal). This balance changes when the strong-weak correlation is intended to change.
        \item Sample a vector of length $n_\emph{task}$, balanced between $0$s and $1$s, representing whether $(m,n)$ or $(r,d)$ should be the more \emph{numerous} integers in the sequence (so that there is no bias, even randomly, between features by size). 
        \item For $i \in 1:n_\emph{task}$:
        \begin{enumerate}
            \item Place the \emph{index} in the first element of an empty array, and the \emph{indicator} in the second.
            \item Based on the $i^{th}$ elements of the three vectors described above, allocate samples of the integers in $(m,n,r,d)_\emph{index}$ into the remaining 18 slots. 
            \item If there are any remaining slots after these integers are randomly allocated, fill them with i.i.d. samples from $unif(1,100)$.
        \end{enumerate}
    \end{enumerate}
\end{enumerate}

\end{document}